\documentclass[10pt,twocolumn,letterpaper]{article}

\usepackage{cvpr}              %

\usepackage{comment}
\usepackage{xcolor}
\usepackage{microtype}
\usepackage{makecell}
\usepackage{soul}
\usepackage{amsmath}
\usepackage{multirow}
\usepackage{wrapfig}

\newcommand{\megascenesnew}{MegaDepth-X\xspace}
\newcommand{\megascenesnewshort}{MD-X\xspace}
\definecolor{mpllightblue}{RGB}{135,206,235} %
\definecolor{mplsteelblue}{RGB}{70,130,180}  %

\definecolor{HKg4Color}{RGB}{0,102,204}   %
\definecolor{JNmtColor}{RGB}{180,60,60}   %
\definecolor{VnV1Color}{RGB}{0,130,90}    %

\definecolor{cvprblue}{rgb}{0.21,0.49,0.74}
\usepackage[pagebackref,breaklinks,colorlinks,allcolors=cvprblue]{hyperref}
\usepackage{booktabs}
\usepackage{multirow}
\usepackage[ruled,vlined]{algorithm2e}
\usepackage{algpseudocode}
\usepackage{soul}

\def\paperID{32887} %
\def\confName{CVPR}
\def\confYear{2026}

\title{Long-Tail Internet Photo Reconstruction}

\author{Yuan Li$^{1}$
\quad
Yuanbo Xiangli$^{1\dagger}$
\quad
Hadar Averbuch-Elor$^{1}$
\quad
Noah Snavely$^{1}$
\quad
Ruojin Cai$^{2\dagger}$
\vspace{2mm}
\\
\centerline{$^1$Cornell University \quad $^2$Kempner Institute, Harvard University}
}

\begin{document}

\twocolumn[{%
\maketitle
\thispagestyle{empty}
\begin{center}
\centering
    \includegraphics[width=\linewidth, trim=0 0 0 0, clip]{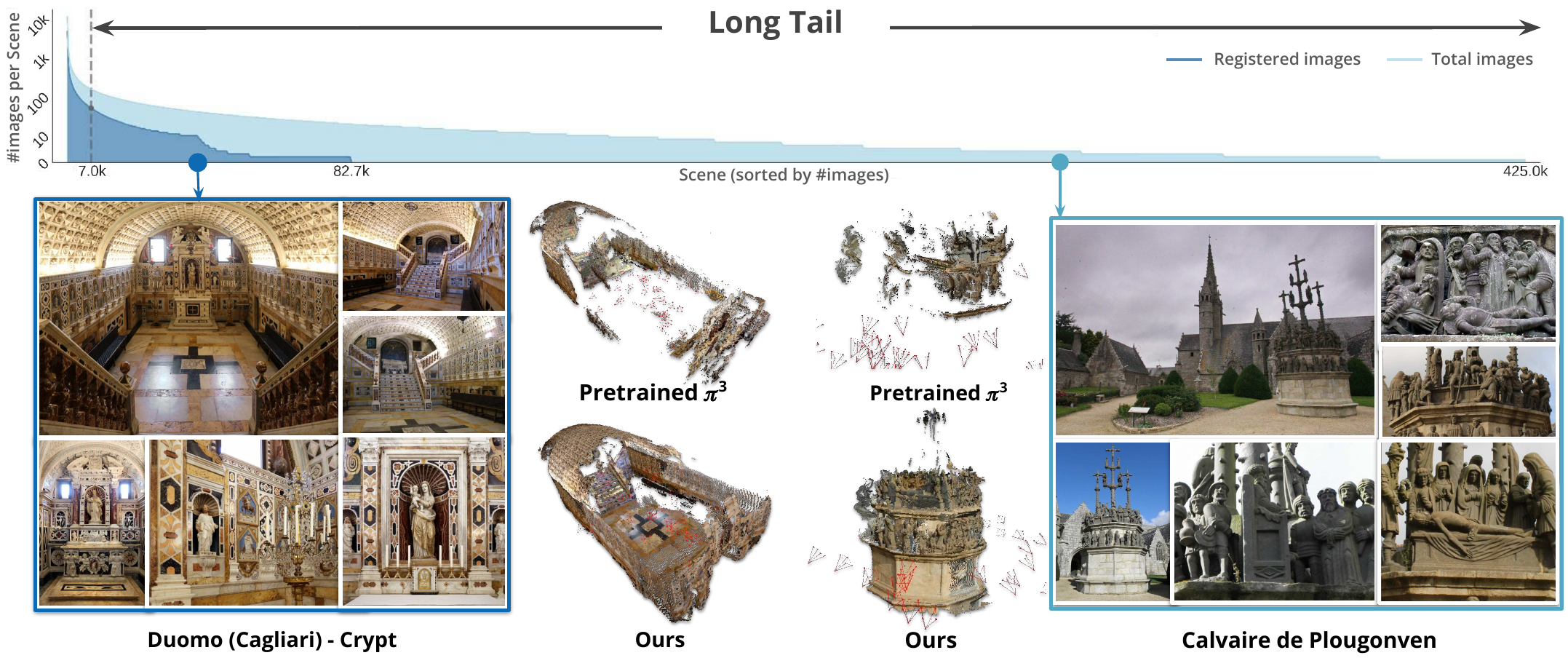}
\vspace{-10pt}
\captionof{figure}{\footnotesize\textbf{Long-tail Internet photo reconstruction.} 
Internet photo collections follow a long-tailed distribution. 
In the top plot, the $x$-axis represents scene index (sorted by image count) and the $y$-axis shows images per scene (scenes are drawn from MegaScenes~\cite{tung2024megascenes}, a dataset of Internet photo collections).
The \textcolor[HTML]{9ddbf0}{light blue} curve plots the total number of Internet photos per scene, while the \textcolor[HTML]{4d98cf}{steel blue} curve shows the size of the subset of photos that were successfully registered using SfM.
The \textit{head} of this distribution of photo collections represents well-photographed scenes; here, there are 6,985 scenes with $>$50 registered images. 
However, most photo collections are in the \textit{long tail} of this distribution; here, 418,056 scenes with fewer than 50 registered photos.
State-of-the-art methods often fail on scenes in this tail.
In the lower half of the figure, 
we show two examples from the long tail, along with representative input images and the corresponding reconstructions.
On Calvaire de Plougonven, COLMAP doesn't register any image;
on both Duomo (Cagliari)-Crypt and Calvaire de Plougonven, recent feed-forward reconstruction models like $\pi^3$~\cite{wang2025pi3} produce poor results. 
We propose \megascenesnew dataset and a strategy for mimicking long-tail camera distributions, on which fine-tuned models like $\pi^3$ exhibit better reconstruction robustness.}
\label{fig:teaser-fig}%
\end{center}%
}]

\def\thefootnote{\fnsymbol{footnote}}
\footnotetext[2]{Corresponding authors.} 
\def\thefootnote{\arabic{footnote}}

\begin{abstract}
Internet photo collections exhibit an extremely long-tailed distribution: a few famous landmarks are densely photographed and easily reconstructed in 3D, while most real-world sites are represented with sparse, noisy, uneven imagery beyond the capabilities of both classical and learned 3D methods. 
We believe that tackling this long-tail regime represents one of the next frontiers for 3D foundation models. 
Although reliable ground-truth 3D supervision from sparse scenes is challenging to acquire, we observe that it can be effectively simulated by sampling sparse subsets from well-reconstructed Internet landmarks.
To this end, we introduce \megascenesnew{}, a large dataset of 3D reconstructions with clean, dense depth,
together with a strategy for sampling sets of training images that mimic camera distributions in long-tail scenes. 
Finetuning 3D foundation models with these components yields robust reconstructions under extreme sparsity,
and also enables more reliable reconstruction in symmetric and repetitive scenes,
while preserving generalization to standard, dense 3D benchmark datasets. 
The dataset, finetuned models, and code are available at: \url{https://megadepth-x.github.io/}. 
\end{abstract}

\section{Introduction}
\label{sec:intro}
Internet photo collections of real-world landmarks follow 
a long-tailed distribution.
A small fraction of famous sites, such as the Colosseum or Notre Dame, are photographed from every conceivable angle and can be accurately reconstructed by standard Structure-from-Motion (SfM) pipelines.
Yet the overwhelming majority of landmarks across the world
are represented on the Internet with just a handful of sparse, noisy
images 
(Fig.~\ref{fig:teaser-fig}).
We refer to this large body of scenes 
as the \emph{long-tail} of online photo collections.
Such scenes are the norm rather than the exception in real-world Internet imagery.

Reconstructing long-tail scenes is challenging. Classic methods, such as COLMAP~\cite{schoenberger2016sfm}, often fail because feature correspondence is hard to find across sparse, non-overlapping, or wide-baseline views. 
Modern learned feed-forward models, like DUSt3R~\cite{wang2023dust3r} and VGGT~\cite{wang2025vggt}, can learn powerful priors from millions of images that might help reconstruct long-tail collections. 
In practice, however, these models are primarily trained on controlled captures with clean, dense, and evenly sampled data. 
When applied to long-tail Internet scenes featuring sparse, diverse, and unevenly distributed imagery, we find that these models often fail to recover consistent geometry.

We believe that one of the next frontiers for 3D foundation models lies in tackling 
this long-tail regime of Internet photos.
Better data is almost certainly key to this problem, 
but we cannot easily construct reliable 3D supervision from long-tail collections themselves, as most contain too few overlapping views for robust reconstruction.
Instead, we propose to \emph{simulate} such long-tailed sets by appropriate sampling of sparse images from  the large, well-reconstructed Internet landmarks at the head of the distribution,
inheriting ground truth from the full reconstruction. 

This strategy requires drawing from large amounts of high-quality landmark reconstructions from Internet photos. 
Existing datasets fall short of this need: MegaDepth~\cite{li2018megadepth} is clean but small, while MegaScenes~\cite{tung2024megascenes} is large but noisy and lacks depth maps. We therefore introduce \textit{\megascenesnew{}}
(dubbed \megascenesnewshort{}), a next-generation extension of MegaDepth in both scale ($7\times$ larger) and quality:
a large-scale, clean, and dense-depth-enhanced dataset built from Internet photo reconstructions with consistent depth refinement and extensive manual verification against reliable references (e.g., Google Maps and satellite imagery). 
Equipped with \megascenesnewshort{}, we propose a novel \emph{sparsity-aware} sampling strategy that mimics the camera distributions of long-tail scenes, encouraging training batches to span wide baselines and partial overlap rather than clustered dense views.

Through extensive experiments, 
we show that models fine-tuned with \megascenesnewshort{} and our sparsity-aware data sampling scheme are significantly more robust on long-tail Internet photo collections, including challenging doppelganger scenes with ambiguous or symmetric content, such as the Calvaire de Plougonven example in Fig.~\ref{fig:teaser-fig}, where classical SfM and pretrained foundation models often fail. In summary, our contributions are:
\begin{itemize}
    \item \textbf{Defining the 3D long-tail regime}: we formalize and characterize the long-tail distribution of Internet photo collections, highlighting this setting's distinct challenges.
    \item \textbf{\megascenesnew{}}, dubbed \megascenesnewshort{}, a large-scale, clean, and depth-augmented dataset for finetuning 3D foundation models on real-world Internet scenes.
    \item \textbf{Sparsity-aware sampling} strategies that simulate the distribution of long-tail Internet collections to improve generalization of 3D prediction models on real-world data.
\end{itemize}

\section{Related Work}

\noindent \textbf{Feed-forward 3D reconstruction.}
Reconstructing 3D scene geometry from 2D images is a cornerstone of computer vision. 
Traditional structure from motion (SfM)~\cite{schonberger2016structure} and multi-view stereo (MVS)~\cite{Schnberger2016PixelwiseVS} methods were crowning achievements of the classic era of 3D vision, and were scaled to large Internet photo collections~\cite{snavely2006photo,agarwal2011building,frahm2010building}.
Recently, the new paradigm of feed-forward 3D reconstruction has emerged, which involves regressing 3D attributes directly from images in a single pass.
Pioneering work in this area, such as DUSt3R, showed success at predicting pixel-aligned point maps from image pairs~\cite{wang2023dust3r}.
MASt3R extended this approach but still relied on pairwise processing~\cite{leroy2024groundingimagematching3d}.
Subsequent efforts focused on scaling these models to arbitrary numbers of views. 
VGGT~\cite{wang2025vggt}, along with concurrent models like Fast3R~\cite{yang2025fast3r} and FLARE~\cite{zhang2025flare}, introduced large transformer architectures that can process hundreds of views simultaneously. 
By leveraging large-scale, diverse datasets and a multi-task learning objective, VGGT predicts a full suite of 3D attributes, including camera parameters, depth maps, and point maps.
To eliminate reference-frame bias, $\pi^{3}$~\cite{wang2025pi3} recently proposed a permutation-equivariant architecture that predicts affine-invariant camera poses and scale-invariant local point maps.
ZipMap~\cite{jin2026zipmap} and Scal3R~\cite{xie2026scal3rscalabletesttimetraining} introduced test-time training approaches to process large image collections.
These methods 
work well on densely-captured and well-conditioned scenes. 
However, we find that their performance on more sparse and noisy Internet photos remains suboptimal, particularly for long-tail scenes.

\medskip \noindent \textbf{Long-tail challenges in 3D vision.}
Long-tailed problems are pervasive in computer vision. They occur when data for common scenarios (the head) are abundant, but examples of rare yet collectively frequent cases (the tail) are scarce. For instance, many object recognition problems involve a few dominant categories but many rarely seen ones, and in autonomous driving, routine driving scenes are plentiful while safety-critical events are hard to capture.

Recently, MegaScenes~\cite{tung2024megascenes} introduced a large-scale scene-level dataset built from Internet photo collections, where long-tail effects are particularly pronounced. Many scenes in the dataset are either unreconstructed or incorrectly reconstructed.  
These failures stem from a combination of view sparsity, noisy imagery, and doppelganger issues~\cite{cai2023doppelgangers}.
Recent work has sought to address such challenges by developing stronger local features~\cite{detone2018superpoint,tyszkiewicz2020disk} and matchers~\cite{Jiang2024OmniGlueGF,sarlin2020superglue,Karpur2023LFM3DLF,lindenberger2023lightglue}, and by learning wide-baseline pose relationships from large-scale 3D datasets~\cite{cai2021extreme,bezalel2025extreme}. The doppelganger problem was further addressed by Cai et al.~\cite{cai2023doppelgangers,xiangli2025doppelgangersimprovedvisualdisambiguation}, who trained classifiers to prune false matches during the structure-from-motion phase of reconstruction.

While these advances have led to enhanced robustness, they do not yet work reliably at scale. 
Ideally, we'd mine ground truth 3D training data for long tail scenes and learn to reconstruct them, but that involves a chicken-and-egg problem,
because the common practice of using available reconstructors (e.g. COLMAP~\cite{schoenberger2016sfm,schoenberger2016mvs}, VGGT~\cite{wang2025vggt}) to derive pseudo-ground-truth camera poses and point maps from natural data doesn't work.
Instead, similar in spirit to approaches used in autonomous driving that augment training data by simulating rare events, our key idea is to take large, well-conditioned image collections and subsample them to simulate long-tailed photo collections, 
and use these to better balance training scene distributions for regression models in order to generalize to long-tailed scenes.

\section{The \megascenesnew{} Dataset}
Learning in the long-tail regime requires high-quality 3D supervision derived from Internet photo collections. This involves two key challenges. First, reconstructions of Internet photo collections can be unreliable due to noise, dynamic content, and ambiguities~\cite{cai2023doppelgangers}.
Second, most long-tail scenes lack any usable reconstructions, as classical SfM pipelines like COLMAP~\cite{schonberger2016structure} often fail on sparse or widely varying image sets. To address these issues, we construct 
\megascenesnewshort{}, 
a large-scale, clean, and depth-refined dataset that provides reliable 3D supervision, built from well-reconstructed scenes in MegaScenes~\cite{tung2024megascenes}.

\begin{figure}[t]
  \centering
  \vspace{-10pt}
   \includegraphics[width=1\linewidth]{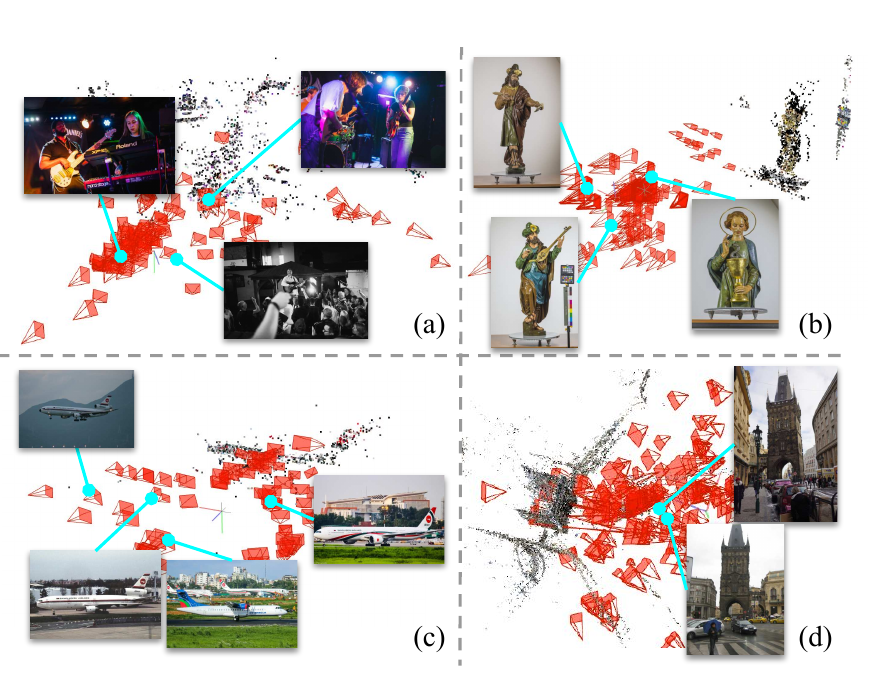}
   \vspace{-20pt}
   \caption{\footnotesize\textbf{Unreliable reconstructions in MegaScenes.} 
   Reconstructions are unreliable when feature matches are incorrectly established on salient, non-static objects (e.g., (a) humans, (b) statues, (c) airplanes) instead of the static scene structure. This results in fragmented and geometrically inconsistent point clouds.
   Example (d) illustrates a doppelganger failure, where images from opposite sides of the building are incorrectly registered together. }
   \label{fig:unreliable_recon}
   \vspace{-4mm}
\end{figure}

\subsection{Filtering and Disambiguation}
\label{sec:filtering}
Our first step in constructing 
\megascenesnewshort{}
is to identify candidate Internet landmarks from which reliable supervision can be derived. We take as our starting pool the subset of MegaScenes with more than 100 registered images, which typically yields stable reconstructions. However, even these ``well-reconstructed'' scenes exhibit two common failure modes: (1) Many scenes contain dynamic events or crowded activities, causing feature matches to lock onto moving objects rather than static structures, leading to unreliable reconstructions. (2) The Doppelganger problem~\cite{cai2023doppelgangers,xiangli2025doppelgangersimprovedvisualdisambiguation}, where visually similar but geographically distant images are mistakenly registered together. Both issues produce incorrect camera poses and fragmented, inconsistent point clouds as shown in Fig.~\ref{fig:unreliable_recon}.

To mitigate these issues, we first inspect the dataset and exclude scenes dominated by crowds or moving objects. Next, we address the doppelganger problem by replacing the default COLMAP SfM reconstruction with MASt3R-SfM~\cite{leroy2024groundingimagematching3d}, combined with Doppelganger classification~\cite{xiangli2025doppelgangersimprovedvisualdisambiguation}. 
Specifically, MASt3R-SfM constructs the scene graph using feature matches derived from MASt3R descriptors, after which the Doppelganger classifier identifies and prunes suspicious edges that may result from doppelganger-induced false correspondences. Finally, we manually verify the reconstructed scenes against external references such as Google Maps and satellite imagery, discarding any scenes that do not align with the corresponding bird’s-eye view.

\subsection{Dense Depth Refinement}
\label{sec:depth_refine}

After obtaining reliable sparse reconstructions, we seek to generate dense depth maps for supervision. We start by running a standard multi-view stereo (MVS)~\cite{schoenberger2016mvs} pipeline.
We observe, as in prior work~\cite{li2018megadepth}, that the resulting geometric depth maps from in-the-wild collections often exhibit artifacts, including depth-bleeding effects (background depths leak into foreground regions) and inconsistent and noisy depths in areas with transient objects (e.g., people, cars).

\begin{figure}[t]
  \centering
   \includegraphics[width=\linewidth]{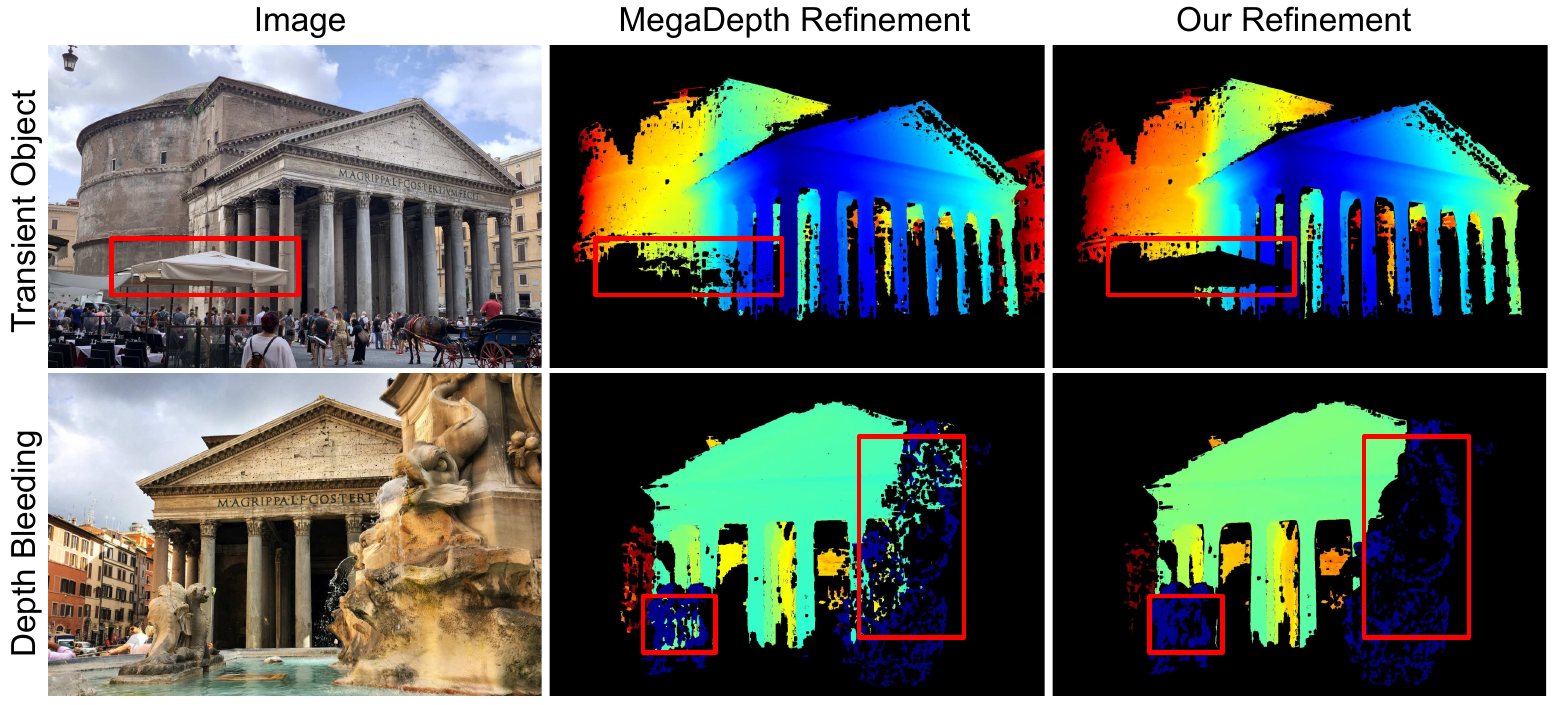}
   \vspace{-20pt}
   \caption{\footnotesize\textbf{Depth refinement.} MVS depth maps often suffer from artifacts like noise from transient objects (top row) and depth bleeding (bottom row). 
   As shown in the middle column, the MegaDepth refinement pipeline (modified MVS, stability filtering, and semantic filtering) fails to fully remedy these issues. Our method (right column) introduces an additional monocular depth-guided filtering step, which effectively removes transient objects and significantly mitigates depth-bleeding artifacts.}
   \label{fig:depth_refinement}
   \vspace{-10pt}
\end{figure}

To address these initial issues, we apply the full depth refinement strategy from MegaDepth~\cite{li2018megadepth},
including a modified MVS procedure that conservatively retains the minimum depth value during propagation, 
stability filtering to remove flickering pixels, 
and semantic filtering to exclude transient objects. 
However, even after this pipeline, we still observe artifacts in the processed geometric depth maps: 
(1) the MegaDepth-modified MVS 
still leads to depth-bleeding artifacts, and (2) semantic filtering is not ideal as it relies on a manually designated list of object categories. 
Examples of such issues are shown in Fig.~\ref{fig:depth_refinement}. 

Therefore, 
to augment MegaDepth's depth refinement procedure, 
we propose
a monocular depth-guided filtering step.
We use depth predictions from MoGe2~\cite{wang2025moge} as ordinal depth priors,
and remove pixels in the processed geometric depth maps that are inconsistent with these priors.
Specifically, we first align the processed geometric depths $D_\text{geom}$ to the monocular predictions $D_\text{mono}$ by matching their median values over valid pixels: $
\small
D'_\text{geom}(p) = s \cdot D_\text{geom}(p), \text{ where }s = \frac{\text{med}\{D_\text{mono}(p) | p \in P\}}{\text{med}\{D_\text{geom}(p) | p \in P\}}
$.
After scale alignment, we compute the normalized depth discrepancy between the two maps:
$
    \Delta(p) = \frac{\lvert{D'_\text{geom}(p)}-{D_\text{mono}(p)}\rvert}{D'_\text{geom}(p)}
$,
and discard pixels whose discrepancies exceed a predefined threshold $\tau_{\text{depth}}$. Moreover, to leverage $D_\text{mono}$ for edge-aware filtering, we compute the discrepancies between the gradients of the two maps: 
$
\Delta(p_{\text{grad}}) = \lvert \frac{\lvert \nabla D_\text{mono} \rvert}{D_\text{mono}} - \frac{\lvert \nabla D'_\text{geom} \rvert}{D'_\text{geom}} \rvert
$ and discard pixels whose discrepancies exceed a predefined threshold $\tau_{\text{grad}}$.
This approach effectively filters both bleeding artifacts and noisy transient objects without relying on manual category lists, as depicted in Fig.~\ref{fig:depth_refinement}.

\subsection{Dataset Statistics}

\label{sec:statistics}
In summary, we identify 2,474 candidate scenes from MegaScenes with more than 100 registered images. Of these, 609 scenes are filtered out due to dynamic content, reconstruction errors, or geometric inconsistencies. Our final 
\megascenesnewshort{}
dataset comprises 1,865 reconstructions totaling 440k images. We reserve 127 scenes for testing, providing a novel set for evaluating both pretrained and fine-tuned methods.
A comparison table with MegaDepth is provided in the supplementary.

\section{Simulating Long-Tail Scenes}

With 
\megascenesnewshort{}
providing reliable 3D supervision, 
the remaining challenge is a complementary supervision coverage problem: existing 3D foundation models are trained predominantly on the head of the Internet-photo distribution, where image collections are large, redundant, and visually well-connected. 
In this regime, models can rely on strong covisibility and abundant local correspondences. However, most real Internet photo collections lie in the long tail, where views are sparse, unevenly distributed, and only weakly connected. 
A more complete 3D prior should therefore be robust not only to diverse scene content, but also to this underrepresented observation regime. 
Rather than seeking unreliable supervision from true long-tail scenes, we start from well-reconstructed scenes in 
\megascenesnewshort{}
and sample subsets whose covisibility structure matches that of real long-tail collections. In this way, we expose the model to the missing part of the training distribution while inheriting trustworthy 3D supervision from the full reconstruction.

\subsection{Defining Properties of Long-Tail Scenes}

Common issues like transient occluders and motion blur affect Internet photos broadly, but they are not the primary bottleneck for long-tail scenes. The more fundamental challenge lies in their viewpoint distribution. In these scenes, sparse camera placements lead to limited mutual overlap between images. This results in fragmented, weakly connected clusters rather than a cohesive set, which poses a major hurdle for reliable 3D reconstruction.
Because accurate camera poses are often unavailable for such scenes, 
we characterize this regime using statistics of the SfM view graph rather than absolute camera geometry.
Our analysis reveals two consistent patterns: (1) \textit{sparser connectivity}: scenes with low registration rates (e.g., only 20\% of images registered) contain a substantially larger fraction of low-degree nodes, with 8\% of cameras having degree two or less, compared with only 3\% in well-reconstructed head scenes. This
indicates that cameras in long-tail scenes are poorly connected, forming fragmented clusters with limited covisibility. (2) \textit{weaker connections}: even among connected image pairs, the average number of geometrically verified feature matches is significantly lower in long-tail scenes than in head scenes (294.8 vs.\ 395.3), indicating reduced overlap and weaker geometric consistency.\footnote{To avoid statistics being dominated by severely noisy scenes, we compute these measurements only on long-tail subsets containing at least five registered images.}
Together, these observations show that the long tail is not simply a regime of fewer images, but one of sparse and weakly connected observation graphs.

Based on these findings, our sampling process should 
satisfy three requirements:
\begin{itemize}
    \item \textbf{Viewpoint Diversity:} The sampled views should cover a wide range of viewing directions, ensuring that emulated scenes span diverse visual perspectives.
    \item \textbf{Sparsity:} The selected views should be far enough apart to mimic the wide baselines typical of long-tail scenes, \eg  loosely connected views or views from disconnected scene components, encouraging the model to learn robust geometric priors rather than relying on dense feature matches.
    \item \textbf{Local Reconstructability:} Despite the sparsity, views within each sampled scene component should retain enough covisibility to remain locally reconstructable,
    since zero-overlap samples within a scene component can lead to unstable training signals and difficult optimization.
\end{itemize}

\subsection{Sparsity-Aware Sampling Strategy}

\begin{figure}[t]
    \centering
    \includegraphics[width=1\linewidth]{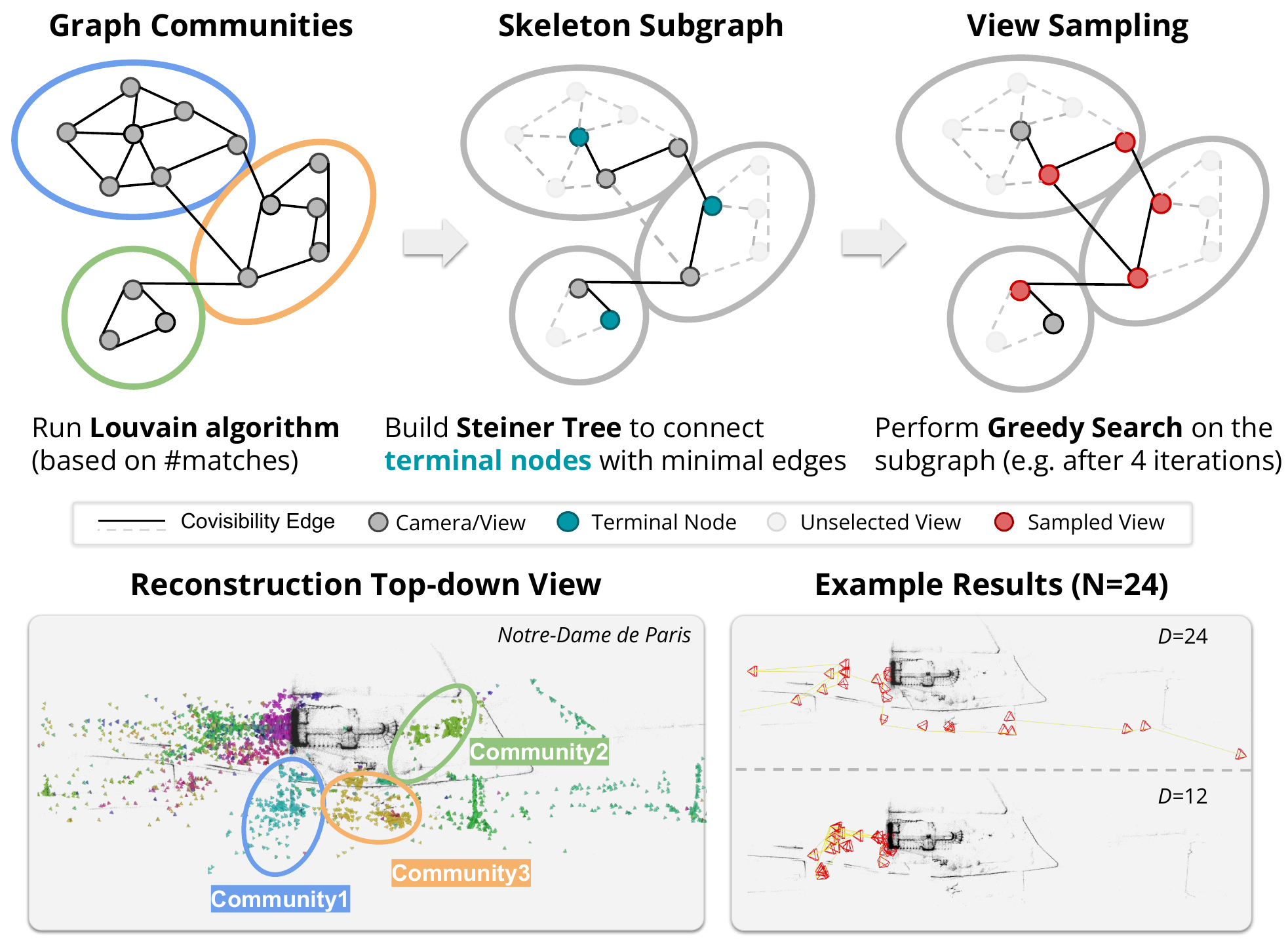}
    \vspace{-20pt}
    \caption{\footnotesize\textbf{Sparsity-aware sampling strategy.} 
\textbf{Top:} Our method follows a multi-stage process: (1) Apply the \textit{Louvain algorithm} to the view graph to identify distinct viewpoint communities. (2) From each community, randomly select a terminal view and construct an approximate \textit{Steiner Tree} to form a minimal, connected subgraph spanning these communities. (3) Perform a \textit{Greedy Search} on this subgraph to select a sparse and diverse set of views. This procedure aims to cover as many communities as possible while ensuring a wide spatial distribution of cameras within each community.
\textbf{Bottom:} A \textit{search depth} parameter controls the final view coverage. In this example, we sample $N=24$ views from the scene with $N_{cc}=1$. With search depth $D=24$, all views are selected via greedy search, producing a more evenly spread distribution. With $D=12$, 12 views come from greedy search and the remaining 12 are sampled locally from the neighborhoods of selected nodes, resulting in a more concentrated distribution.
}
    \label{fig:sampling_pipeline}
    \vspace{-3mm}
\end{figure}

We therefore formulate the sampling task as sampling $N$ views that form at most $N_{cc}$ connected components,
in order to emulate a long-tail scene 
with multiple weakly connected or disconnected scene components.
Specifically, components are allowed to be disconnected from one another, but within each sampled component we still require sufficient internal covisibility for local reconstructability.
We find that naïve random or uniform subsampling 
often fails to satisfy this balance,
producing either 
zero-overlap sets within scene components or clusters biased toward dense regions.
We instead propose a structured sampling process. 
We first partition views into strongly connected communities and then select a minimal yet diverse subset that ensures both community coverage and global connectivity.
This process is illustrated in Fig.~\ref{fig:sampling_pipeline}.

\medskip\noindent\textbf{Graph Communities.}
To promote viewpoint diversity in our sampling, we first identify the dominant ``viewing areas” within each scene. We represent the SfM structure as a view graph $G = (V, E)$, where each node $v_i \in V$ corresponds to a camera view and each edge $(v_i, v_j) \in E$ is weighted by the number of feature matches $w_{ij}$. We prune edges with $w_{ij} < 50$ to remove minor overlaps, resulting in a filtered graph $G' = (V, E')$ that preserves only meaningful covisibility relationships.
To reveal clusters of cameras with dense internal connectivity, we perform community detection (e.g., Louvain community detection~\cite{blondel2008fast}) on the view graph. This yields viewpoint groups ${C_k}$ that efficiently capture distinct visual regions and the dominant perspectives of the scene. We then randomly partition the graph into $N_{cc}$ connected components that span different communities and do the following steps \emph{within each graph partition}. The partition algorithm is provided in the supplementary material.

\medskip\noindent\textbf{Minimal Connectivity Subgraph.}
To preserve overall scene connectivity while maintaining sparsity and view diversity within limited nodes, we construct a minimal structure linking all identified communities without reintroducing dense redundancy within each partition. 
We then compute an approximate Steiner tree to link all of these nodes \cite{kou1981fast, mehlhorn1988faster}.\footnote{A Steiner tree aims to span a specified set of \emph{terminal} nodes while introducing only the minimal set of intermediate nodes required for connectivity.}
In particular, for each training batch for a given training scene, we first randomly select one representative view $v_k \in C_k$ from each community $C_k$ to form the terminal set $T = \{v_k\}$. An approximate Steiner tree algorithm then constructs a minimal connected subgraph $G_\text{sub} = (V_\text{sub}, E_\text{sub}), \quad T \subseteq V_\text{sub} \subseteq V$, that spans all terminal nodes using only the necessary intermediate nodes. This yields a compact subgraph connecting all communities using the fewest necessary nodes and edges, preserving global consistency while retaining sparsity.
Since $G_\text{sub}$ can have an arbitrary number of nodes, we need to perform additional sampling to get desired number of views for the training and testing batches.

\medskip\noindent\textbf{Greedy View Sampling.}
Inspired by 
skeletal sets~\cite{snavely2008skeletal}, we perform greedy view sampling on the subgraph $G_\text{sub}$ to select a diverse subset of views for long-tail emulation. 
The objective is to iteratively expand the sampled set toward 
broad spatial coverage
while maintaining sufficient 
covisibility
among selected view pairs. 

At each iteration, the algorithm 
aims to select the next view based on two criteria:
(1) \emph{Community novelty}: preferring cameras that belong to previously unseen communities, thereby introducing new viewing directions and reducing redundancy; and
(2) \emph{Spatial distance}: encouraging selection of cameras farther from the current viewpoint to promote wider baseline coverage.
Specifically,
the algorithm operates on a current node $v$ and its connected neighborhood $N_v$.
Let $S$ denote the set of already sampled nodes and $M$ be the community map. %
We first determine which communities have already been reached in $S$, forming the set
$S_{\text{comm}} = \{ M[s] \mid s \in S \}$.
For each neighbor $u \in N_v$, we then evaluate its community novelty by checking whether $M[u] \notin S_{\text{comm}}$, and compute its spatial distance as $\| \mathrm{Pos}(u) - \mathrm{Pos}(v) \|_2$, where $\mathrm{Pos}(\cdot)$ is camera position. 
Details for this algorithm are provided in the supplemental material.
All candidate neighbors are ranked lexicographically by these two attributes, and the top-ranked neighbor $u^*$ is selected as the next sampled node. 
This procedure is repeated for $D$ iterations (i.e., the search depth).

\medskip\noindent\textbf{Implementation.} In practice, we compute a fixed set of communities $\mathcal{C} = \{C_k\}$ for each scene. 
To form a training batch of $N$ images for a scene, we first randomly divide the $N$ samples across all $N_{cc}$ partitions. 
In each partition, greedy view sampling stops once either a predefined search-depth limit $D$
is reached or the target number of views assigned to that partition has been sampled. 
Here, $D$ controls how far the search expands within a partition, hence the sparsity of the resulting set.
If this process still produces fewer than $N$ nodes in total, we fill the remaining slots by randomly sampling nodes from the local neighborhoods of the previously sampled nodes.
Fig.~\ref{fig:sampling_pipeline} illustrates an example in which $N=24$ and $N_{cc}=1$,
and shows the different sparsities of the sampled set obtained under different values of $D$.
Before training, we run the proposed sampling algorithm offline to generate mini-batches of 24 nodes, avoiding costly graph loading during training. We then perform depth-first search from random seed nodes to subsample 2 to 24 images for training batches.

\section{Experiments}
We evaluate how our approach improves 3D reconstruction in the long-tail regime of Internet photo collections. 
First, we show quantitative results on the proposed 
\megascenesnewshort{}
benchmark, demonstrating qualitative improvements on real-world long-tail and doppelganger scenes. We then analyze the effect of the proposed dataset and sampling strategy, and finally verify that our fine-tuned models preserve strong performance on standard, curated benchmarks. 
Further implementation details and additional results are in the supplementary material. 

\subsection{Experimental Setup}
\label{subsec:exp_setup}
\noindent\textbf{Backbones and variants.}
We finetune two feed-forward 3D foundation models, $\pi^3$~\cite{wang2025pi3} and VGGT~\cite{wang2025vggt}, on 
\megascenesnewshort{}
using our proposed sampling strategy. 
We adopt the loss functions from $\pi^3$~\cite{wang2025pi3} and VGGT~\cite{wang2025vggt}. To preserve pretrained geometric fidelity, we finetune only the Alternating-Attention modules and keep the point cloud and camera decoders frozen. More training details are in the supplementary. The resulting models are denoted as $\pi^3$-\textsc{FT} and VGGT-\textsc{FT}.

To study how our proposed view sampling strategy affects performance, we finetune $\pi^3$ on clean Internet data using four sampling schemes:
\begin{itemize}
  \item \textsc{Dense}: training batches with densely overlapping views where $D=5$ and $N_{cc} = 1$,
  \item \textsc{Sparse}: long-tail--like sampling emphasizing wide baselines where $D=24$ and $N_{cc} = 4$,
  \item \textsc{Mixed}: a combination of dense and sparse batches for balanced learning with $D\in[5, 24]$ and $N_{cc} \in [1, 4]$,
  \item \textsc{Random}: random view sampling.
\end{itemize}
Unless otherwise noted, \textsc{FT} (e.g., $\pi^3$-\textsc{FT}) refers to the model finetuned on the cleaned dataset using the \textsc{Mixed} sampling strategy above.
We additionally train a \textsc{Dirty} variant on Internet data (using the same Mixed scheme) without the filtering strategy in Sec.~\ref{sec:filtering}, while keeping the same depth refinement pipeline in Sec.~\ref{sec:depth_refine}, to assess robustness to label noise and data contamination.

\medskip\noindent\textbf{Evaluation Metrics.}
For camera pose estimation, we follow prior work~\cite{wang2025vggt,wang2025pi3} and report
Relative Rotation Accuracy (RRA), Relative Translation Accuracy (RTA), and their combined Area Under Curve (AUC).
We also report mean rotation and translation errors (MRE and MTE, in degrees).  
For point map evaluation, we follow prior work~\cite{azinovic2022nrgbd,wang2023dust3r,wang2024spann3r,wang2025cut3r,wang2025pi3} and report Accuracy (Acc), Completeness (Comp), and Normal Consistency (NC), each computed as the mean and median across test scenes.

\subsection{Internet Photo Evaluation}
\label{subsec:internet_photo_eval}
We first evaluate models on the proposed 
\megascenesnewshort{}
benchmark, which contains Internet photo collections of varying sparsity and difficulty.
For each test scene, we sample 24 images from the reconstructed scene graph using our sampling algorithm, and categorize them into 
\emph{easy} ($D=5$, $N_{cc}=1$) and \emph{hard} ($D=24$, $N_{cc}=4$)
subsets according to the greedy search depth used for test data sampling.

\begin{table}[t]
\caption{\footnotesize\textbf{Quantitative results on \megascenesnew{}} for camera pose and point map estimation across two difficulty levels.
 Our finetuned models ($\pi^3$-\textsc{FT} and VGGT-\textsc{FT}) trained with the proposed dataset and sampling strategy consistently outperform pretrained baselines, especially on harder, sparser scenes.}
\vspace{-2mm}
\label{tab:MegaScenes_results}
\centering
\footnotesize
\renewcommand{\arraystretch}{0.9}
\setlength{\tabcolsep}{3pt}
\resizebox{1\linewidth}{!}{
    \begin{tabular}{llccccccccccc}
    \toprule
    && \multicolumn{5}{c}{\textbf{Camera Pose Estimation}}& \multicolumn{6}{c}{\textbf{Point Map Estimation}}\\
    \cmidrule(lr){3-7} \cmidrule(lr){8-13}
    & \multirow{2}{*}[-0.3em]{\textbf{Method}} 
    & \multirow{2}{*}[-0.3em]{RRA@5$\uparrow$}
    & \multirow{2}{*}[-0.3em]{RTA@5$\uparrow$}
    & \multirow{2}{*}[-0.3em]{AUC@5$\uparrow$}
    & \multirow{2}{*}[-0.3em]{MRE$\downarrow$}
    & \multirow{2}{*}[-0.3em]{MTE$\downarrow$}
    & \multicolumn{2}{c}{Acc$\downarrow$}
    & \multicolumn{2}{c}{Comp$\downarrow$}
    & \multicolumn{2}{c}{NC$\uparrow$} \\
    \cmidrule(lr){8-9} \cmidrule(lr){10-11} \cmidrule(lr){12-13}
    & &&&&& & Mean & Med. & Mean & Med. & Mean & Med. \\
    \midrule
     \multirow{4}{*}[-0.3em]{\rotatebox[origin=c]{90}{\emph{easy}}}    
     &$\pi^3$ & 88.97 & 68.79 & 45.84 & 4.12 & 7.82 & 0.055 & 0.030 & 0.039 & 0.019 & 0.712 & 0.822 \\				
    
     &$\pi^3$-\textsc{FT} & \textbf{95.64} & \textbf{76.85} & \textbf{55.58} & \textbf{1.64} & \textbf{5.50} & \textbf{0.035} & \textbf{0.020} & \textbf{0.024} & \textbf{0.012} & \textbf{0.724} & \textbf{0.837} \\ \cmidrule(lr){2-13}
    
     &VGGT & 84.17 & 58.47 & 35.32 & 4.55 & 9.93 & 0.093 & 0.047 & 0.055 & 0.026 & 0.695 & 0.798 \\				
     
     &VGGT-\textsc{FT} & \textbf{92.41} & \textbf{71.12} & \textbf{48.78} & \textbf{2.70} & \textbf{7.02} & \textbf{0.050} & \textbf{0.027} & \textbf{0.033} & \textbf{0.014} & \textbf{0.719} & \textbf{0.833} \\
    \midrule
     \multirow{4}{*}[-0.3em]{\rotatebox[origin=c]{90}{\emph{hard}}} 
     &$\pi^3$ & 75.31 & 59.16 & 36.93 & 12.21 & 10.82 & 0.101 & 0.065 & 0.133 & 0.090 & 0.689 & 0.786 \\	
    
     &$\pi^3$-\textsc{FT} & \textbf{86.40} & \textbf{71.00} & \textbf{47.93} & \textbf{5.72} & \textbf{7.27} & \textbf{0.068} & \textbf{0.041} & \textbf{0.066} & \textbf{0.041} & \textbf{0.713} & \textbf{0.818} \\	\cmidrule(lr){2-13}			
    
     &VGGT & 70.98 & 52.98 & 29.10 & 13.20 & 13.34 & 0.149 & 0.092 & 0.151 & 0.104 & 0.675 & 0.764 \\				
     
     &VGGT-\textsc{FT} & \textbf{81.07} & \textbf{65.59} & \textbf{41.49} & \textbf{7.22} & \textbf{9.05} & \textbf{0.089} & \textbf{0.053} & \textbf{0.084} & \textbf{0.055} & \textbf{0.709} & \textbf{0.814} \\
    \bottomrule
    \end{tabular}
    }
\vspace{-3mm}
\end{table}

\medskip\noindent\textbf{Quantitative Results.}
Tab.~\ref{tab:MegaScenes_results} reports quantitative results for camera pose and point map estimation across three difficulty levels on 
\megascenesnewshort{}.
Finetuning markedly improves both $\pi^3$ and VGGT over their pretrained baselines, with larger gains observed in harder, sparser scenes. 
These improvements hold across metrics
indicate that the fine-tuned models better capture global structure and maintain consistent 3D geometry in sparse settings.

\begin{table}[t]
\centering
\caption{\footnotesize\textbf{Ablation study on \megascenesnew{}.}
 Finetuning on the cleaned dataset with \textsc{Mixed} dense–sparse sampling ($\pi^3$-\textsc{FT}) yields the best overall performance, while training on unfiltered data (\textsc{Dirty}) degrades accuracy.
 }
\vspace{-2mm} 
\label{tab:MegaScenes_ablation}
\footnotesize
\renewcommand{\arraystretch}{0.9}
\setlength{\tabcolsep}{3pt}
\resizebox{1\linewidth}{!}{
     \begin{tabular}{llccccccccccc}
    \toprule
    && \multicolumn{5}{c}{\textbf{Camera Pose Estimation}}& \multicolumn{6}{c}{\textbf{Point Map Estimation}}\\
    \cmidrule(lr){3-7} \cmidrule(lr){8-13}
    & \multirow{2}{*}[-0.3em]{\textbf{Method}} 
    & \multirow{2}{*}[-0.3em]{RRA@5$\uparrow$}
    & \multirow{2}{*}[-0.3em]{RTA@5$\uparrow$}
    & \multirow{2}{*}[-0.3em]{AUC@5$\uparrow$}
    & \multirow{2}{*}[-0.3em]{MRE$\downarrow$}
    & \multirow{2}{*}[-0.3em]{MTE$\downarrow$}
    & \multicolumn{2}{c}{Acc$\downarrow$}
    & \multicolumn{2}{c}{Comp$\downarrow$}
    & \multicolumn{2}{c}{NC$\uparrow$} \\
    \cmidrule(lr){8-9} \cmidrule(lr){10-11} \cmidrule(lr){12-13}
    & &&&&& & Mean & Med. & Mean & Med. & Mean & Med. \\
    \midrule
     \multirow{6}{*}{\rotatebox[origin=c]{90}{\emph{easy}}}    
     &$\pi^3$ & 88.97 & 68.79 & 45.84 & 4.12 & 7.82 & 0.055 & 0.030 & 0.039 &	0.019 &	0.712 &	0.822 \\
     &$\pi^3$-\textsc{FT} & \underline{95.64} & \textbf{76.85} & \underline{55.58} & \underline{1.64} & \textbf{5.50} & \textbf{0.035} & \textbf{0.020} &	\textbf{0.024} &	\textbf{0.012} &	\underline{0.724} &	\textbf{0.837} \\
     &$\pi^3$-\textsc{Dirty} & 91.25 & 72.80 & 51.77 & 5.16 & 7.28 & 0.075 &	0.052 &	0.081 &	0.051 &	0.710 &	0.818 \\
     &$\pi^3$-\textsc{Random} & 95.08 & 76.42 & 55.00 & 1.78 & 5.72 & 0.039 & 0.021 & \underline{0.026} &	\underline{0.013} &	0.720 &	0.831 \\				
     &$\pi^3$-\textsc{Dense} & 95.13 & \underline{76.73} & \textbf{55.65} & 1.84 & 5.61 & \underline{0.036} &	\textbf{0.020} &	 \underline{0.026} &	\underline{0.013} &	\textbf{0.725} &	\textbf{0.837} \\		
     &$\pi^3$-\textsc{Sparse} & \textbf{96.27} & 76.46 & 55.12 & \textbf{1.61} & \underline{5.59} & 0.038 & \textbf{0.020} &  \underline{0.026} & \underline{0.013} & 0.723 & \underline{0.835} \\	
    \midrule
     \multirow{6}{*}{\rotatebox[origin=c]{90}{\emph{hard}}} 
     &$\pi^3$ & 75.31 & 59.16 & 36.93 & 12.21 & 10.82 & 0.101 &	0.065 &	0.133 &	0.090 &	0.689 &	0.786 \\
     &$\pi^3$-\textsc{FT} & \textbf{86.40} & \textbf{71.00} & \textbf{47.93} & \textbf{5.72} & \textbf{7.27} & \textbf{0.068} &	\underline{0.041} &	\underline{0.066} &	\underline{0.041} &	\textbf{0.713} &	\textbf{0.818} \\
     &$\pi^3$-\textsc{Dirty} & 81.10 & 65.99 & 43.74 & 11.86 & 9.72 & 0.130 &	0.094 &	0.139 &	0.091 &	0.693 &	0.791 \\
     &$\pi^3$-\textsc{Random} & 85.93 & 69.84 & 47.17 & 6.53 & 7.78 & 0.071 &	\textbf{0.040} &	0.073 &	0.045 &	0.708 &	0.812 \\
     &$\pi^3$-\textsc{Dense} & 85.82 & 70.06 & \underline{47.47} & \underline{6.04} & 7.64 & 0.071 &	0.042 &	\textbf{0.062} &	\textbf{0.035} &	\textbf{0.713} &	\underline{0.817} \\			
     &$\pi^3$-\textsc{Sparse} & \underline{85.97} & \underline{70.53} & 47.13 & 6.05 & \underline{7.52} & \underline{0.070} & \textbf{0.040} & 0.070 & \underline{0.041} & \underline{0.710} & 0.814 \\		
     \bottomrule
    \end{tabular}
    }
\vspace{-3mm}
\end{table}

\medskip\noindent\textbf{Ablation Analysis.}
We analyze the effects of data quality and sampling strategies, with results shown in Tab.~\ref{tab:MegaScenes_ablation}. 
Training on unfiltered (\textsc{Dirty}) data consistently reduces accuracy, even performing worse than the pretrained model in point-map estimation on both the \emph{easy} and \emph{hard} levels, highlighting the importance of clean supervision for robust generalization. 
Among sampling schemes, \textsc{Random} sampling yields reasonable camera pose accuracy but provides limited improvement in point map reconstruction, emphasizing the importance of adequate covisibilities in training batches.
\textsc{Dense} sampling performs well on easier scenes but is less effective under sparse conditions. 
\textsc{Sparse} sampling alone does not yield the best trade-off. Although it exposes the model to more challenging cases, \textsc{Mixed} sampling achieves slightly better overall performance across difficulty levels.

\medskip\noindent\textbf{Qualitative Analysis.} We show qualitative results for three settings: the 
\megascenesnewshort{}
test set, real-world long-tail Internet scenes, and doppelganger scenes.

\begin{figure}[t]
    \centering
    \includegraphics[width=0.95\linewidth]{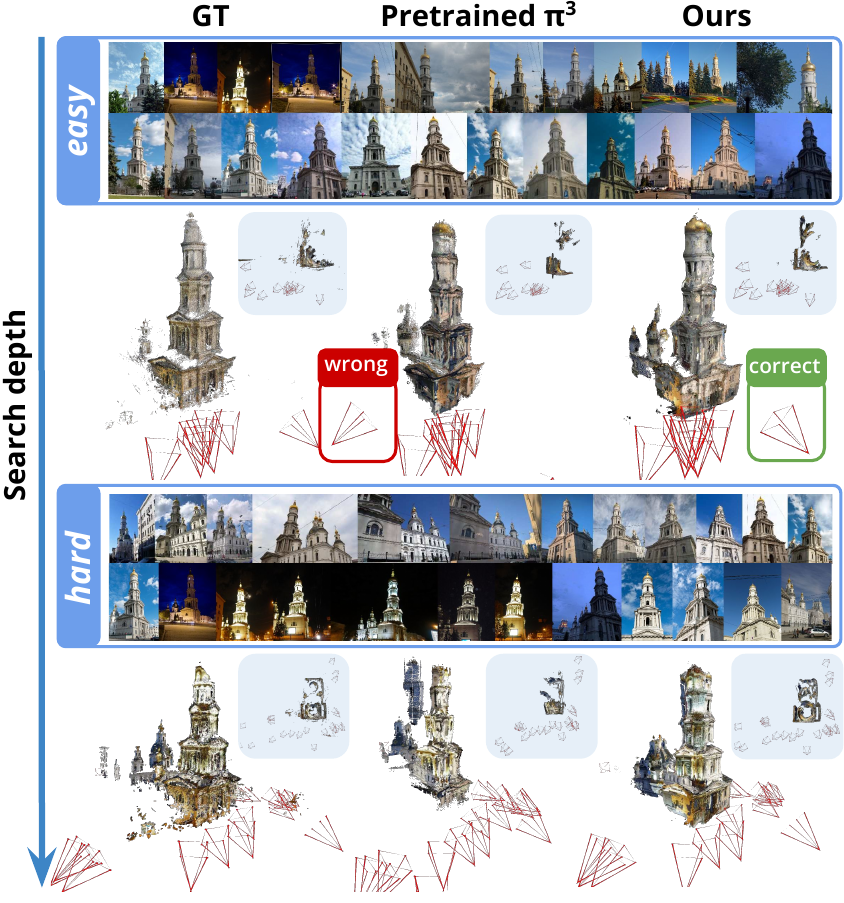}
    \vspace{-10pt}
    \caption{\footnotesize\textbf{Reconstruction results on the \megascenesnew{} test set across two difficulty levels.} 
    For each level, the top row shows the full 24-image input set, and the bottom row compares reconstructions from ground truth, pretrained $\pi^3$, and our finetuned model with top-down views shown in the insets. 
    Our model shows clearer improvements in the \textit{hard} setting, where the inputs are more challenging.
    Note that \textit{hard} was obtained using a deeper search depth than \textit{easy.}}

    \label{fig:dataset_level_vis}
\end{figure}

\noindent\textit{\megascenesnew{} Visualization.}
Fig.~\ref{fig:dataset_level_vis} shows reconstruction results on the 
\megascenesnewshort{}
test set across \emph{easy} and \emph{hard} levels. 
Our fine-tuned model produces more accurate camera poses, more dense and consistent 3D point maps compared to the pretrained baseline, especially on sparse (\emph{hard}) scenes. 
It generalizes well across varying camera intrinsics and challenging appearance changes such as day-night shifts. 

\begin{figure*}[t]
    \centering
    \includegraphics[width=0.98\linewidth]{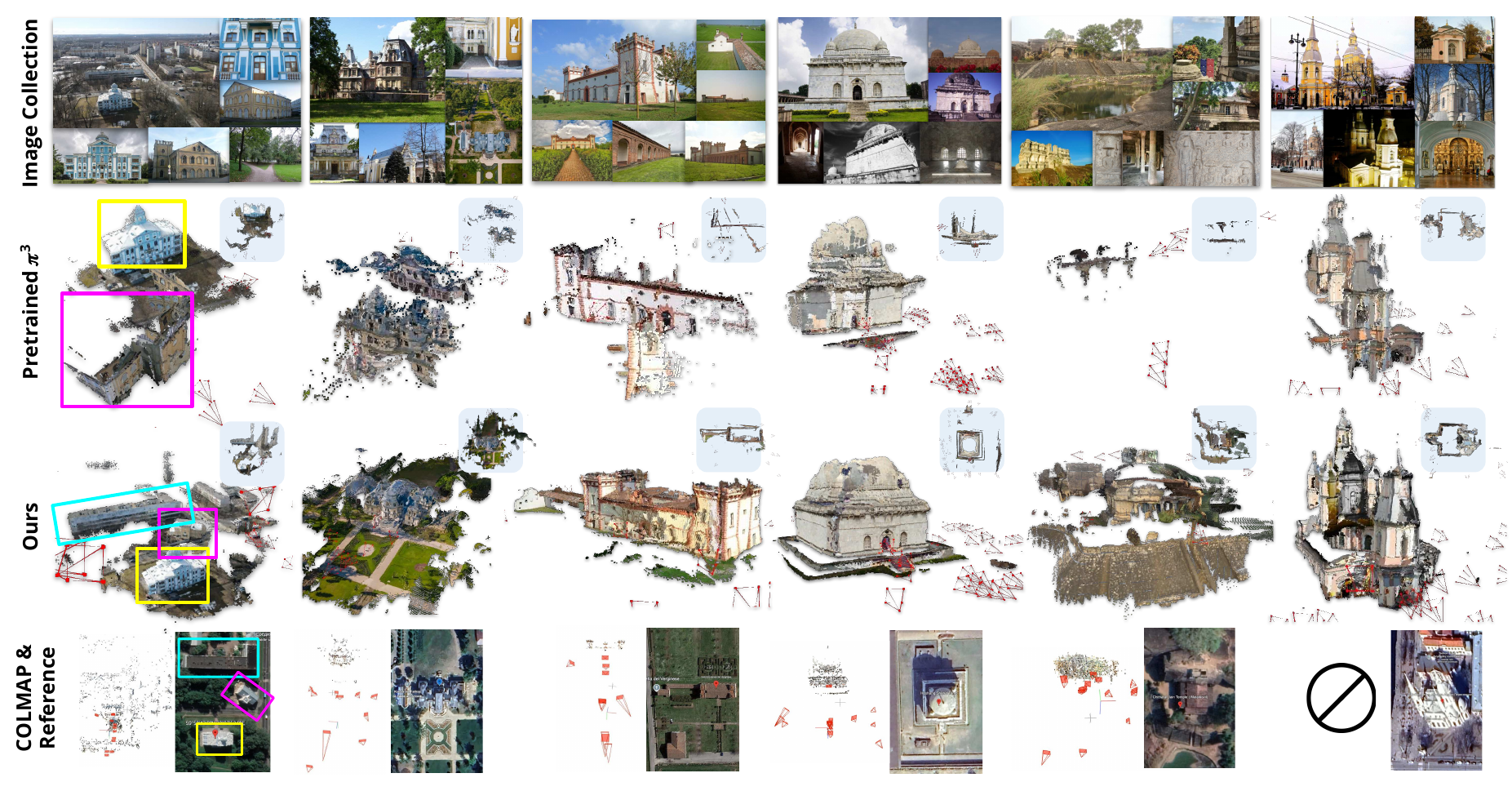}
    \vspace{-10pt}
    \caption{\footnotesize\textbf{Reconstruction results on real long-tail Internet scenes.}
    Each scene contains only a handful of photos with uneven viewpoints and noisy content, where COLMAP fails to register most images and produces extremely sparse geometry. 
    Pretrained $\pi^3$ makes low-confidence predictions and incomplete reconstructions,
    while our fine-tuned model 
    discovers the correct large-scale layout (e.g., (1) \textit{Novo-Znamenka Manor}, 66 images, 13 registered), handles very few-view inputs and recovers dense geometry ((2) \textit{Sobanski Palace in Guzow}, 95 images, 11 registered), 
    reconstructs more complete structures under sparse, long-tail settings ((3) \textit{Delizia del Verginese (Gambulaga, Portomaggiore)}, 69 images, 11 registered, (5) \textit{Chitharal Jain Monuments}, 44 images, 15 registered), 
    resolves doppelganger ambiguity ((4) \textit{Hoshang's Tomb}, 85 images, 40 registered),
    and even works when COLMAP completely fails ((6) \textit{Chapel of Saint Andrew's cathedral (Saint Petersburg)}, 94 images, 0 registered).
    These results demonstrate that our model remains robust and confident under severe sparsity and ambiguity in real long-tail Internet scenes. \textbf{For each scene, the confidence threshold is the same for pretrained $\pi^3$ and our method.}
    }
    \label{fig:longtail_preview}
    \vspace{-3mm}
\end{figure*}

\noindent\textit{Real Long-Tail Scenes.}
Real long-tail Internet scenes often contain fewer than 100 usable photos captured from uneven viewpoints and mixed with transient or irrelevant content.
Classical SfM pipelines, e.g., COLMAP, typically fail to register most images, producing extremely sparse geometry or incomplete reconstructions.
Pretrained models struggle under these conditions, yielding low-confidence predictions and fragmented structures. 
Our finetuned model remains stable and reconstructs coherent global geometry. 
As shown in Fig.~\ref{fig:longtail_preview}, our model successfully reconstructs dense geometry from very few views, and handles doppelganger ambiguities with higher confidence, demonstrating strong robustness and generalization to real-world long-tail scenes. In the supplementary material, we provide more results on doppelganger scenes.

\subsection{Generalization to Standard Benchmarks}
We next examine whether the finetuned models preserve generalization on standard, curated benchmarks.

\begin{table}[t]
\caption{\footnotesize\textbf{Camera pose estimation on RealEstate10K~\cite{zhou2018stereo} and CO3Dv2~\cite{Reizenstein2021CommonOI}}. We follow $\pi^3$'s pose sampling conventions. 
Our fine-tuned models, trained on proposed Internet data dataset, remain comparable to pretrained baselines, demonstrating generalization to standard benchmarks.}
\vspace{-2mm}
\label{tab:RE10K_CO3Dv2_relpose}
\centering
\footnotesize
\renewcommand{\arraystretch}{0.95}
\renewcommand{\arraystretch}{0.95}
\setlength{\tabcolsep}{3.5pt}
\resizebox{1\linewidth}{!}{
    \begin{tabular}{lcccccccccc}
    \toprule
  & \multicolumn{5}{c}{\textbf{RealEstate10K}} & \multicolumn{5}{c}{\textbf{CO3Dv2}}\\
  \cmidrule(lr){2-6} \cmidrule(lr){7-11}
     \textbf{Method}
    & {RRA@5$\uparrow$}
    & {RTA@5$\uparrow$}
    & {AUC@5$\uparrow$}
    & {MRE$\downarrow$}
    & {MTE$\downarrow$}  
    & {RRA@5$\uparrow$}
    & {RTA@5$\uparrow$}
    & {AUC@5$\uparrow$}
    & {MRE$\downarrow$}
    & {MTE$\downarrow$}\\
    \midrule
     $\pi^3$
    & 98.79 & \textbf{79.61} & \textbf{62.82} & \textbf{0.51} & \textbf{5.65}  & 93.24 & 84.47 & 57.12 & 3.04 & 4.28 \\
     $\pi^3$-\textsc{FT}
    & \textbf{98.80} & 77.78 & 60.01 & \textbf{0.51} & 6.13  & \textbf{93.97} & \textbf{84.50} & \textbf{57.61} & \textbf{2.96} &\textbf{4.26}  \\ 
    \cmidrule(lr){1-11}
    VGGT
    & 97.49 & 62.32 & 38.09 & 1.03 & 8.66  & 96.97 & 86.19 & \textbf{67.84} & 2.33 &3.95  \\
     VGGT-\textsc{FT}
    & \textbf{98.23} & \textbf{71.88} & \textbf{48.23}  & \textbf{0.82} & \textbf{6.85}  & \textbf{97.11} & \textbf{86.27} & 67.81 & \textbf{2.29} &\textbf{3.92}  \\
    \bottomrule
    \end{tabular}
    }

\end{table}

\smallskip \noindent\textbf{Relative Pose Estimation.}
We evaluate on RealEstate-10K~\cite{zhou2018stereo} and CO3Dv2~\cite{Reizenstein2021CommonOI}, following $\pi^3$’s pose sampling conventions. 
As shown in Tab.~\ref{tab:RE10K_CO3Dv2_relpose}, fine-tuning on Internet data generally maintains the performance of both backbones, and yields modest improvements for VGGT in particular. These results indicate that robustness learned from sparse, in-the-wild Internet photos does not compromise generalization to standard 3D benchmarks.

\begin{table}[t]
\caption{\footnotesize\textbf{Point map estimation on DTU~\cite{jensen2014large} and ETH3D~\cite{schops2017multi}}. 
Finetuning on the proposed Internet photo dataset retain overall reconstruction quality on DTU, while performance on ETH3D decreases due to domain mismatch with Internet imagery. These results show that the model adapts to Internet photos without drifting too much on out-of-domain benchmarks.}
\vspace{-2mm}
\label{tab:dtu_eth3d}
\centering
\footnotesize
\renewcommand{\arraystretch}{0.95}
\setlength{\tabcolsep}{3.5pt}
\resizebox{1.05\linewidth}{!}{
\begin{tabular}{lcccccccccccc}
\toprule
 & \multicolumn{6}{c}{\textbf{DTU}} & \multicolumn{6}{c}{\textbf{ETH3D}} \\
\cmidrule(lr){2-7} \cmidrule(lr){8-13}
\textbf{Method} &
\multicolumn{2}{c}{Acc. $\downarrow$} &
\multicolumn{2}{c}{Comp. $\downarrow$} &
\multicolumn{2}{c}{N.C. $\uparrow$} &
\multicolumn{2}{c}{Acc. $\downarrow$} &
\multicolumn{2}{c}{Comp. $\downarrow$} &
\multicolumn{2}{c}{N.C. $\uparrow$} \\
\cmidrule(lr){2-3} \cmidrule(lr){4-5} \cmidrule(lr){6-7}
\cmidrule(lr){8-9} \cmidrule(lr){10-11} \cmidrule(lr){12-13}
 & Mean & Med. & Mean & Med. & Mean & Med. & Mean & Med. & Mean & Med. & Mean & Med. \\
\midrule
$\pi^3$ 
& \textbf{1.151} & \textbf{0.622} & \textbf{1.793} & 0.629 & \textbf{0.668} & \textbf{0.754}
& \textbf{0.188} & \textbf{0.126} & \textbf{0.211} & \textbf{0.129} & \textbf{0.872} & \textbf{0.967} \\
$\pi^3$-\textsc{FT} 
& 1.202 & 0.642 & 1.928 & \textbf{0.593} & 0.666 & 0.751
& 0.199 & 0.142 & 0.242 & 0.151 & 0.861 & 0.955 \\
\cmidrule(lr){1-13}
VGGT 
& 1.308 & 0.761 & 1.929 & 1.015 & 0.665 & 0.750 
& \textbf{0.270} & \textbf{0.174} & \textbf{0.304} & \textbf{0.180} & \textbf{0.841} & \textbf{0.942} \\
VGGT-\textsc{FT} 
& \textbf{1.283} & \textbf{0.759} & \textbf{1.900} & \textbf{0.953} & \textbf{0.669} & \textbf{0.756} 
& 0.282 & 0.205 & 0.394 & 0.225 & 0.838 & 0.927 \\
\bottomrule
\end{tabular}
}
\end{table}

\begin{table}[t]
\caption{\footnotesize\textbf{Point map estimation on 7-Scenes~\cite{shotton2013scene} and NRGBD~\cite{azinovic2022nrgbd} datasets.} 
We evaluate both sparse-view and dense-view settings. 
Finetuning on Internet photos yields comparable performance to pretrained baselines with minor variations, indicating our method preserves generalization across diverse real world and synthetic datasets.}
\vspace{-2mm}
\label{tab:NRGBD_7Scenes_comparison}
\centering
\footnotesize
\renewcommand{\arraystretch}{0.9}
\setlength{\tabcolsep}{3.5pt}
\resizebox{1.05\linewidth}{!}{
\begin{tabular}{llccccccccccccc}
\toprule
 & & & \multicolumn{6}{c}{\textbf{7-Scenes}} & \multicolumn{6}{c}{\textbf{NRGBD}} \\
\cmidrule(lr){4-9} \cmidrule(lr){10-15}
\textbf{View} & \textbf{Method} & 
& \multicolumn{2}{c}{Acc. $\downarrow$}
& \multicolumn{2}{c}{Comp. $\downarrow$}
& \multicolumn{2}{c}{NC. $\uparrow$}
& \multicolumn{2}{c}{Acc. $\downarrow$}
& \multicolumn{2}{c}{Comp. $\downarrow$}
& \multicolumn{2}{c}{NC. $\uparrow$} \\
\cmidrule(lr){4-5} \cmidrule(lr){6-7} \cmidrule(lr){8-9}
\cmidrule(lr){10-11} \cmidrule(lr){12-13} \cmidrule(lr){14-15}
 & & & Mean & Med. & Mean & Med. & Mean & Med. & Mean & Med. & Mean & Med. & Mean & Med. \\
\midrule
\multirow{4}{*}{\textit{sparse}}
& $\pi^3$ &  & 0.047 & 0.029 & 0.074 & 0.049 & \textbf{0.741} & 0.840 
& \textbf{0.024} & \textbf{0.013} & \textbf{0.028} & \textbf{0.013} & \textbf{0.909} & \textbf{0.991} \\
 & $\pi^3$-\textsc{FT}
& & \textbf{0.046} & \textbf{0.027} & \textbf{0.072} & \textbf{0.046} & 0.739 & \textbf{0.841} 
& \textbf{0.024} & 0.014 & \textbf{0.028} & 0.014 & 0.903 & 0.990 \\
\cmidrule(lr){2-15}
 & VGGT 
& & \textbf{0.044} & \textbf{0.024} & \textbf{0.056} & \textbf{0.033} & 0.733 & \textbf{0.846} 
& \textbf{0.049} & \textbf{0.027} & \textbf{0.066} & \textbf{0.037} & \textbf{0.882} & \textbf{0.979} \\
& VGGT-\textsc{FT} 
& & 0.062 & 0.046 & 0.097 & 0.070 & \textbf{0.738} & 0.844 
& 0.071 & 0.046 & 0.071 & 0.041 & 0.875 & 0.959 \\
\midrule
\multirow{4}{*}{\textit{dense}}
& $\pi^3$ &  & \textbf{0.016} & \textbf{0.007} & \textbf{0.022} & \textbf{0.011} & \textbf{0.689} & \textbf{0.792} 
& \textbf{0.013} & \textbf{0.007} & \textbf{0.014} & 0.006 & \textbf{0.874} & \textbf{0.981} \\
&$\pi^3$-\textsc{FT} 
& & \textbf{0.016} & \textbf{0.007} & 0.023 & \textbf{0.011} & 0.686 & 0.789 
& \textbf{0.013} & \textbf{0.007} & \textbf{0.014} & \textbf{0.005} & 0.864 & 0.978 \\
\cmidrule(lr){2-15}
&VGGT 
& & 0.022 & 0.008 & \textbf{0.026} & \textbf{0.012} & 0.667 & 0.760 
& \textbf{0.015} & \textbf{0.008} & \textbf{0.015} & \textbf{0.006} & \textbf{0.871} & \textbf{0.982} \\
&VGGT-\textsc{FT} 
& & \textbf{0.016} & \textbf{0.007} & 0.027 & \textbf{0.012} & \textbf{0.681} & \textbf{0.781} 
& \textbf{0.015} & 0.008 & 0.016 & \textbf{0.006} & 0.859 & 0.981 \\
\bottomrule
\end{tabular}
}\
\vspace{-6mm}

\end{table}

\medskip\noindent\textbf{Point Map Estimation.}
Results on DTU~\cite{jensen2014large}, ETH3D~\cite{schops2017multi}, 7-Scenes~\cite{shotton2013scene}, and NRGBD~\cite{azinovic2022nrgbd} (Tab.~\ref{tab:dtu_eth3d}\&\ref{tab:NRGBD_7Scenes_comparison}) show that 
our model maintains comparable reconstruction accuracy
on DTU, 7-Scenes and NRGBD.
We observe a performance decrease on ETH3D and a mild drop for VGGT under sparse NRGBD, likely reflecting the domain gap between these clean, controlled datasets and Internet imagery.
Overall, the results indicate that training on diverse Internet photos preserves cross-dataset generalization without overfitting.

\section{Conclusion}
We presented a step towards robust, Internet-scale 3D reconstruction by defining and addressing the long-tail regime of Internet photo collections. 
Through the \megascenesnew{} dataset and a sparsity-aware sampling strategy, we augment the ability of 3D foundation models to recover consistent geometry from sparse, noisy, and ambiguous imagery, where classical SfM and SOTA feed-forward 3D reconstruction models fail, 
and demonstrates disambiguation of doppelganger scenes
while maintaining generalization across benchmarks.

Our dataset currently focuses on landmark-scale scenes, representing only a small fraction of the landscape of Internet photos. Bootstrapping on the current dataset and refining models for reconstructions of even more longed-tail data remains an important direction for future work.
Extending this framework beyond landmarks to everyday objects, indoor scenes, and other Internet photo domains offers a promising path toward a truly universal 3D foundation model.

\paragraph{Acknowledgments}
This work was supported in part by the Institute of Information \& Communications Technology Planning \& Evaluation (IITP) grant funded by the Korean Government (MSIT) (No. RS-2024-00457882, National AI Research Lab Project).
We thank Joseph Tung, Yiwen Zhang, Hanyu Chen and Haian Jin for discussion and help with MegaScenes dataset and depth post-processing.

{
    \small
    \bibliographystyle{ieeenat_fullname}
    \bibliography{main}
}

\clearpage
\appendix
\setcounter{page}{1}
\maketitlesupplementary

\setcounter{tocdepth}{2}

\section*{Visualization Webpage}
\addcontentsline{toc}{section}{Visualization Webpage}
Please refer to our 
\href{https://megadepth-x.github.io/}{project page}
for additional visualizations beyond this PDF. 
The webpage includes:
(i) animations of our sparsity-aware sampling procedure on representative scenes;
and (ii) comparisons of reconstructions from pretrained $\pi^3$ and our finetuned $\pi^3$ on long-tail scenes (where COLMAP registers 0 images).
We also provide video fly-throughs of reconstructed point clouds and additional qualitative results on the webpage to help visualize performance on diverse, real-world scenes.

\section{The \megascenesnew{} Dataset}

\subsection{Data Processing}
\begin{figure}
    \centering
    \includegraphics[width=1\linewidth]{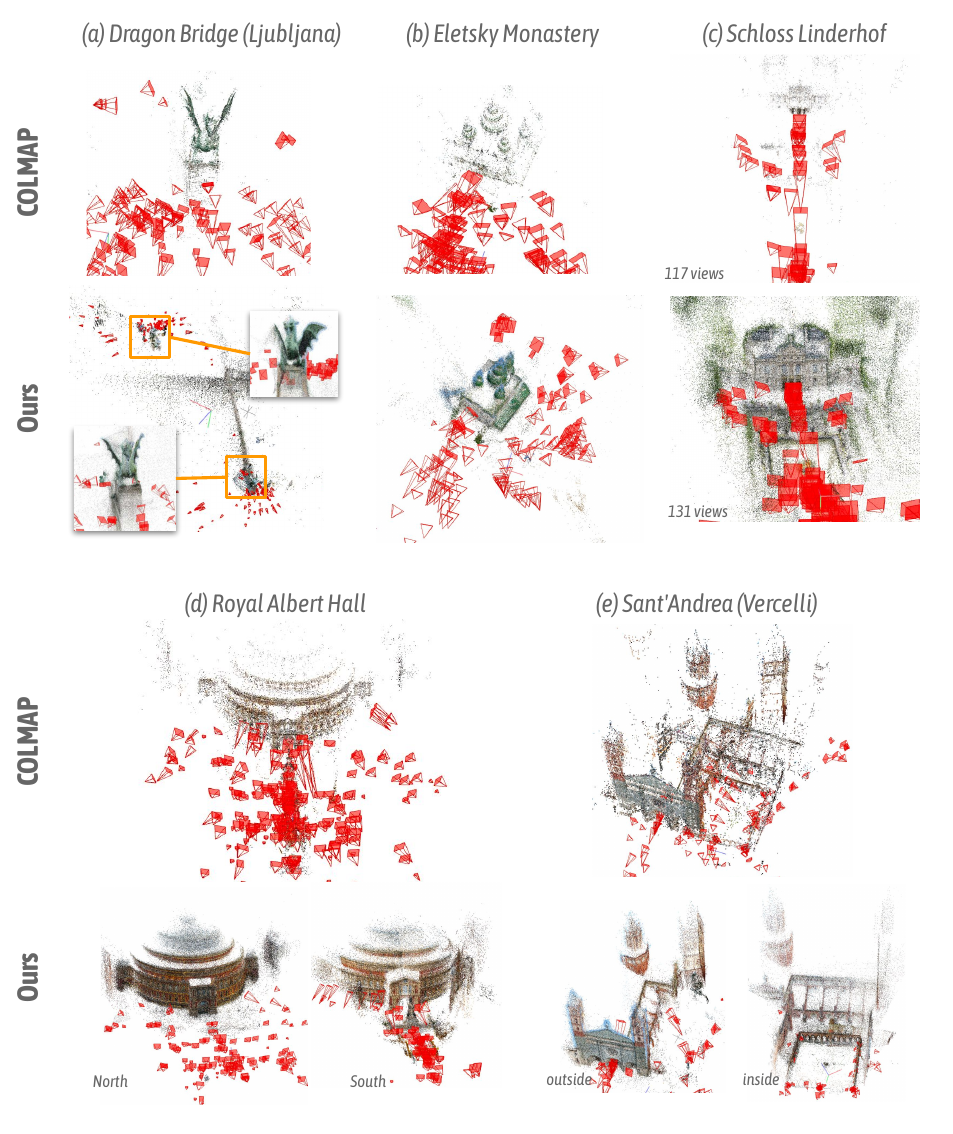}
    \caption{\textbf{Comparison of COLMAP and our reconstruction pipeline.}We replace COLMAP with MASt3R-SfM \cite{duisterhof2025mastrsfm} combined with the doppelganger++ classifier \cite{xiangli2025doppelgangersimprovedvisualdisambiguation} to obtain sparse reconstructions, allowing effective disambiguation of doppelganger scenes.
(a) The bridge has two similar dragon statues, one at each end. COLMAP incorrectly treats them as the same statue and registers them together, whereas our method correctly separates them.
(b), (d), and (e) illustrate additional doppelganger cases, in which different sides or parts of a landmark are mistakenly merged.
(c) In this low-texture scene, our pipeline also succeeds in registering more images.  }
    \label{fig:colmap_vs_ours}
\end{figure}

\begin{figure}
    \centering
    \includegraphics[width=1\linewidth]{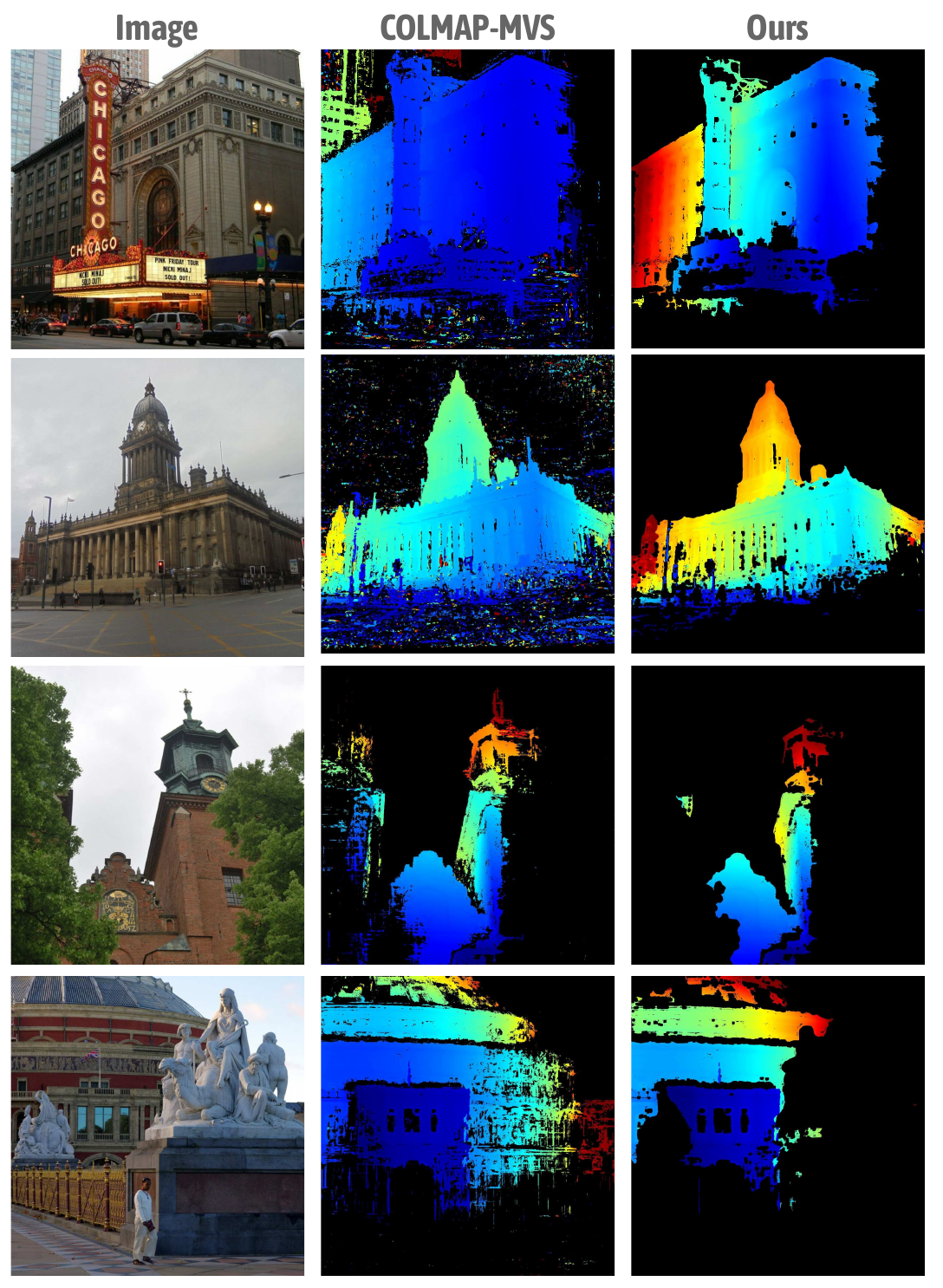}
    \caption{\textbf{Comparison of COLMAP MVS and our filtered dense depth results.} COLMAP MVS suffers from depth bleeding and struggles to correctly estimate the depth of transient objects. Our strategy mitigates these issues by leveraging ordering priors from monocular depth predictions. Note that we prioritize \emph{accurate} depth maps over complete ones. In the last row, COLMAP fails to recover the depth of the foreground statue, and we opt to remove the depth values in that region. If we were to warp the monocular depth to match the MVS result, then any inaccuracy in the relative depth between the statue and the background building could produce erroneous and inconsistent cross-view depth estimates.}
    \label{fig:mvs_compare}
\end{figure}

In this section, we compare COLMAP results with those produced by our proposed data-processing pipeline.
Fig.~\ref{fig:colmap_vs_ours} shows reconstructions from COLMAP and our MASt3R-SfM pipeline. COLMAP often fails on ambiguous scenes involving similar-looking objects, visually similar but distinct building facades, symmetric landmarks etc. In contrast, our reconstruction pipeline effectively mitigates these issues and recovers correct geometry. In Fig.~\ref{fig:mvs_compare}, we show that our monocular depth–guided dense depthmap filtering strategy prevents background depths from leaking into foreground regions (i.e. the depth-bleeding issue~\cite{li2018megadepth}) and removes depth estimates on transient objects, which are often unreliable in COLMAP MVS. Note that we use monocular depth only as guidance, rather than warping it to align with the MVS depth. This is because we prioritize \emph{accurate} depth maps over complete ones. Uncertainty in the relative depth predictions of monocular models can introduce additional noise and inconsistency across views. For example, in the last row of Fig.~\ref{fig:mvs_compare}, COLMAP MVS fails to recover the depth of the foreground statue, and we opt to remove the depth values in that region. If we were to warp the monocular depth to match the MVS result, then any inaccuracy in the relative depth between the statue and the background building could produce erroneous and inconsistent cross-view depth estimates.

\subsection{Dataset Statistics}

We provide an overall comparison between MegaDepth and \megascenesnew{} in Tab.~\ref{tab:dataset_stats}, including reconstruction statistics as well as several metrics that characterize the spatial distribution of viewpoints. Beyond basic dataset properties such as the number of intact reconstructions, image count, and whether doppelganger filtering or dense depth refinement is applied, we analyze how cameras are positioned and oriented in each scene, as scenes with broad viewpoint coverage allow our sampling strategy to construct more diverse and representative sparse-view subsets. The statistics are computed from Manhattan-aligned COLMAP reconstructions.

\paragraph{Positional coverage.}
To understand how cameras are placed in the horizontal plane, we compute each camera's azimuth angle relative to the scene centroid (that is, the angle of the direction from the scene centroid to the camera) and divide the full 0-360° range into 36 equal 10° bins. 
In practice, the scene centroid is derived from the average of the SFM point cloud. 
A scene with many occupied bins is one where cameras are well-distributed around the object. 
In the table, the columns “Positional Azimuth Coverage $=100\%$ / $\ge75\%$ / $\ge50\%$ / $\ge25\%$” report how many scenes achieve at least that percentage of bins($36/36$, $27/36$, $18/36$, $9/36$), with larger thresholds indicating closer to full 360° wrap-around coverage.

\paragraph{Rotational coverage.}
Position alone does not describe where cameras are looking. We therefore measure the coverage of camera orientations by mapping each camera’s forward viewing direction to 36 azimuth bins similar to positional coverage. If cameras face more distinct directions, more bins are occupied; if they face similar directions, only few bins are occupied. We summarize this rotational azimuth coverage using the same percentage thresholds as positional azimuth coverage.

These statistics show that \megascenesnew{} contains substantially more scenes with broad camera-position coverage and diverse viewing directions, making it better suited for robust sparse-view reconstruction than MegaDepth.

\begin{table*}[t]
\centering
\caption{\textbf{Dataset statistics and viewpoint-distribution metrics.}
We report reconstruction statistics and metrics describing camera coverage. 
\textit{Positional Azimuth Coverage} counts scenes whose camera positions occupy 9–36 (i.e. 25\%-100\%) of the 36 horizontal azimuth bins (10° per bin, covering the full 360°). \textit{Rotational Azimuth Coverage} represents scenes whose camera forwarding vectors occupy 9–36 (i.e. 25\%-100\%) of the 36 horizontal azimuth bins (10° per bin, covering the full 360°). For each scene, the more bins covered, the wider the camera distribution is.
\textsuperscript{$\dagger$}Dense depth refinement uses monocular depth–guided filtering.}
\label{tab:dataset_stats}

\setlength{\tabcolsep}{2.5pt}
\resizebox{1\linewidth}{!}{
\begin{tabular}{lcccclcccclcccc}
\toprule
\multirow{2}{*}[-0.3em]{\textbf{Dataset}} & \multirow{2}{*}[-0.3em]{\makecell{\textbf{\#Recons.}}} & \multirow{2}{*}[-0.3em]{\textbf{\#Images}} & \multirow{2}{*}[-0.3em]{\makecell{\textbf{Doppelganger} \\ \textbf{Check}}} & \multirow{2}{*}[-0.3em]{\makecell{\textbf{Dense Depth}\\ \textbf{Refinement}}} & & \multicolumn{4}{c}{\textbf{Positional Azimuth Coverage}}& &\multicolumn{4}{c}{\textbf{Rotational Azimuth Coverage}}\\
 \cmidrule(lr){7-10} \cmidrule(lr){12-15}  
&&&&& & \makecell{ = 100\% $\uparrow$} & \makecell{ $\ge$ 75\% $\uparrow$} & \makecell{ $\ge$ 50\% $\uparrow$} &  \makecell{ $\ge$ 25\% $\uparrow$}  & &\makecell{ = 100\% $\uparrow$} & \makecell{ $\ge$ 75\% $\uparrow$} & \makecell{ $\ge$ 50\% $\uparrow$} &  \makecell{ $\ge$ 25\% $\uparrow$}  \\
\midrule
MegaDepth~\cite{li2018megadepth} & 266 & 119k & No & Yes &   & 4 & 15 & 25 & 74  & & 27 & 56 & 107 & 230 \\
\megascenesnew{} (Ours) & 1,865 & 440k & Yes & \makecell{Yes\textsuperscript{$\dagger$}}& & 6 & 80 & 223 & 752 & & 76 & 490 & 1123 & 1816  \\
\bottomrule
\end{tabular}
}
\end{table*}

\section{Sparsity-aware Sampling}

\subsection{Greedy Sampling Algorithm}
We illustrate one iteration of the greedy view-sampling procedure in Alg.~\ref{alg:one_iter_sampling}.
At each step, the algorithm selects the next view based on two criteria:
\begin{enumerate}
    \item \textit{Community novelty}: prioritizing candidates whose camera-community has not yet been visited by the sampled set. This encourages the trajectory to enter unexplored regions of the view graph and reduces redundancy in viewpoint selection.
    \item \textit{Spatial distance}: among candidates with equal novelty, preferring those that are farther from the current camera position. This promotes larger baselines and helps diversify the spatial coverage of the sampled views.
\end{enumerate}
Candidates are lexicographically ranked according to these two criteria, and the highest-ranked node is chosen as the next sampled view. 

\begin{algorithm}[t]
\label{alg:sampling}
\SetAlgoLined
\caption{One Step of Greedy View Sampling}
\label{alg:one_iter_sampling}

\KwIn{
    Current node $v$ \\
    Neighborhood of $v$: $N_v$ \\
    Set of already sampled nodes $S$ \\
    Community map $M$ (node $\rightarrow$ community) \\
    Camera positions $\mathrm{Pos}(\cdot)$
}
\KwOut{
    Next sampled node $u^*$
}
\BlankLine

\small{\tcp{Identify communities already covered}}
$S_\text{comm} \gets \{ M[s] \mid s \in S \}$\;

\small{\tcp{Compute candidate list with community novelty and distance}}
$\mathcal{C} \gets \emptyset$\;%
\For{\textbf{each} $u \in N_v$}{
    $\textit{unreached} \gets (M[u] \notin S_\text{comm})$\;
    $\textit{dist} \gets \|\mathrm{Pos}(u) - \mathrm{Pos}(v)\|_2$\;
    $\mathcal{C} \gets \mathcal{C} \cup \{(u, \textit{unreached}, \textit{dist})\}$\;
}

\small{\tcp{Sort by unreached, then by distance}}
Sort $\mathcal{C}$ in descending lexicographic order by $(\textit{unreached}, \textit{dist})$\;

\small{\tcp{Select the top-ranked candidate}}
$(u^*, \_, \_) \gets \text{first element of } \mathcal{C}$\;

\Return{$u^*$}\;

\end{algorithm}

\subsection{Graph Partition}

Before sparsity-aware sampling, we partition COLMAP’s view graph into $N_{cc}$ subgraphs. Specifically, we randomly select $N_{cc}$ seed nodes and treat each seed as the initial node of one partition. Starting from these seeds, we perform a parallel round-robin breadth-first search(BFS) over the view graph. During each iteration, every subgraph expands from its current frontier to its unassigned neighboring nodes, which are then incorporated into that subgraph. In this way, each node is assigned to the subgraph of the seed that first reaches it, until no further nodes can be expanded.

\begin{algorithm}[t]
\caption{Round-Robin BFS Graph Partitioning}
\label{alg:graph_partition}
\SetAlgoLined

\KwIn{
    View graph $G = (V, E)$ \\
    Number of subgraphs $N_{cc}$
}
\KwOut{
    Subgraphs $\{\mathcal{P}_1, \dots, \mathcal{P}_{N_{cc}}\}$
}
\BlankLine

Randomly select $N_{cc}$ seed nodes $\{s_1, \dots, s_{N_{cc}}\} \subseteq V$\;
Initialize each $\mathcal{P}_i$ with seed $s_i$\;
Initialize one BFS frontier for each subgraph\;

\While{there exists a non-empty frontier}{
    \For{\textbf{each} subgraph $\mathcal{P}_i$}{
        Expand its frontier by one BFS step\;
        Assign each newly reached unassigned node to $\mathcal{P}_i$\;
    }
}

\Return{$\{\mathcal{P}_1, \dots, \mathcal{P}_{N_{cc}}\}$}\;

\end{algorithm}

\subsection{Graph Span vs.\ Search Depth}
\begin{figure*}[h!]
    \centering
    \begin{subfigure}[t]{0.245\textwidth}
        \centering
        \includegraphics[width=\textwidth,trim={0.2cm 0 0.2cm 1.04cm},clip]{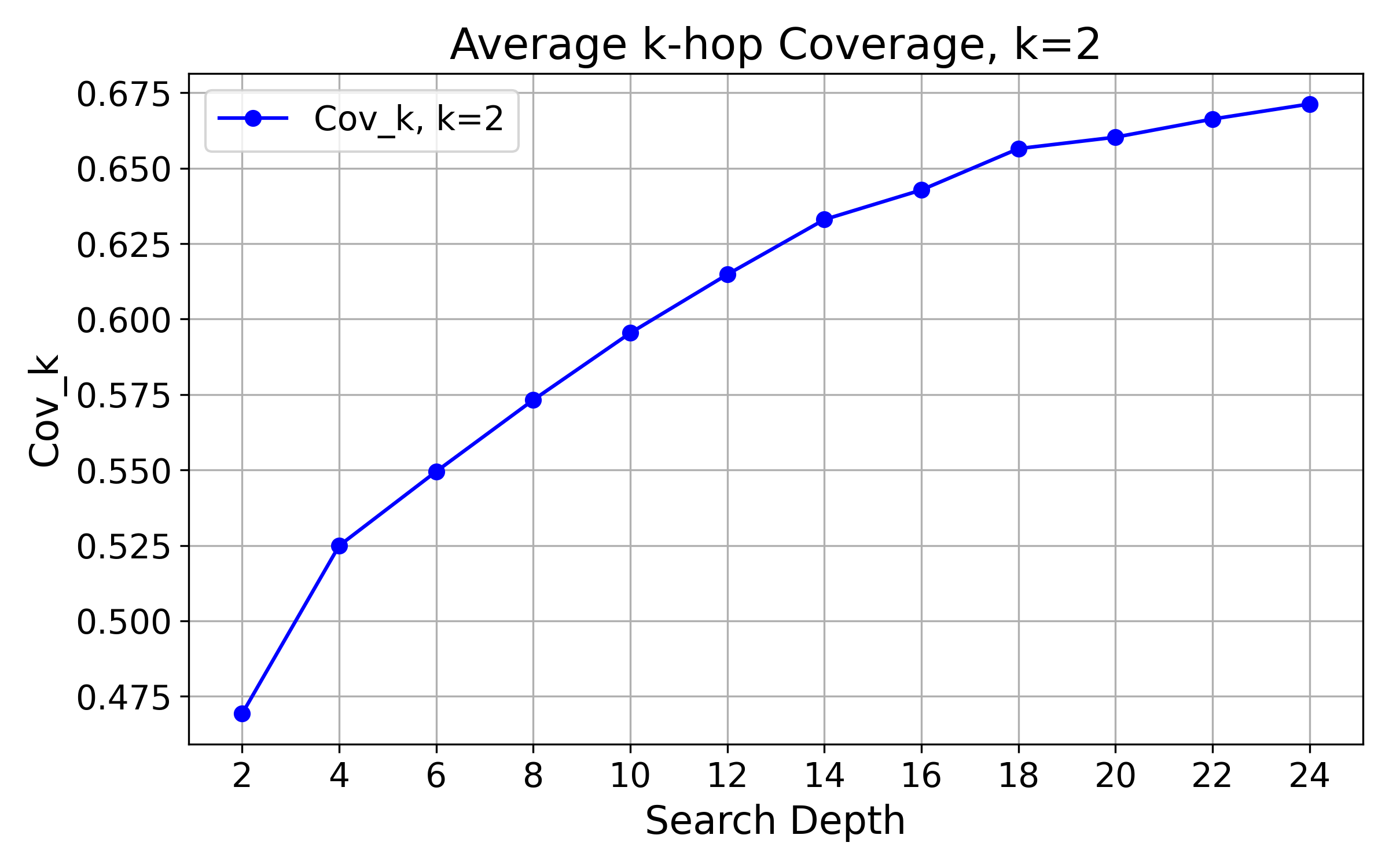}
        \caption{\scriptsize k-hop Coverage ($k{=}2$)}
    \end{subfigure}
    \hfill
    \begin{subfigure}[t]{0.245\textwidth}
        \centering
        \includegraphics[width=\textwidth,trim={0.2cm 0 0.2cm 1.04cm},clip]{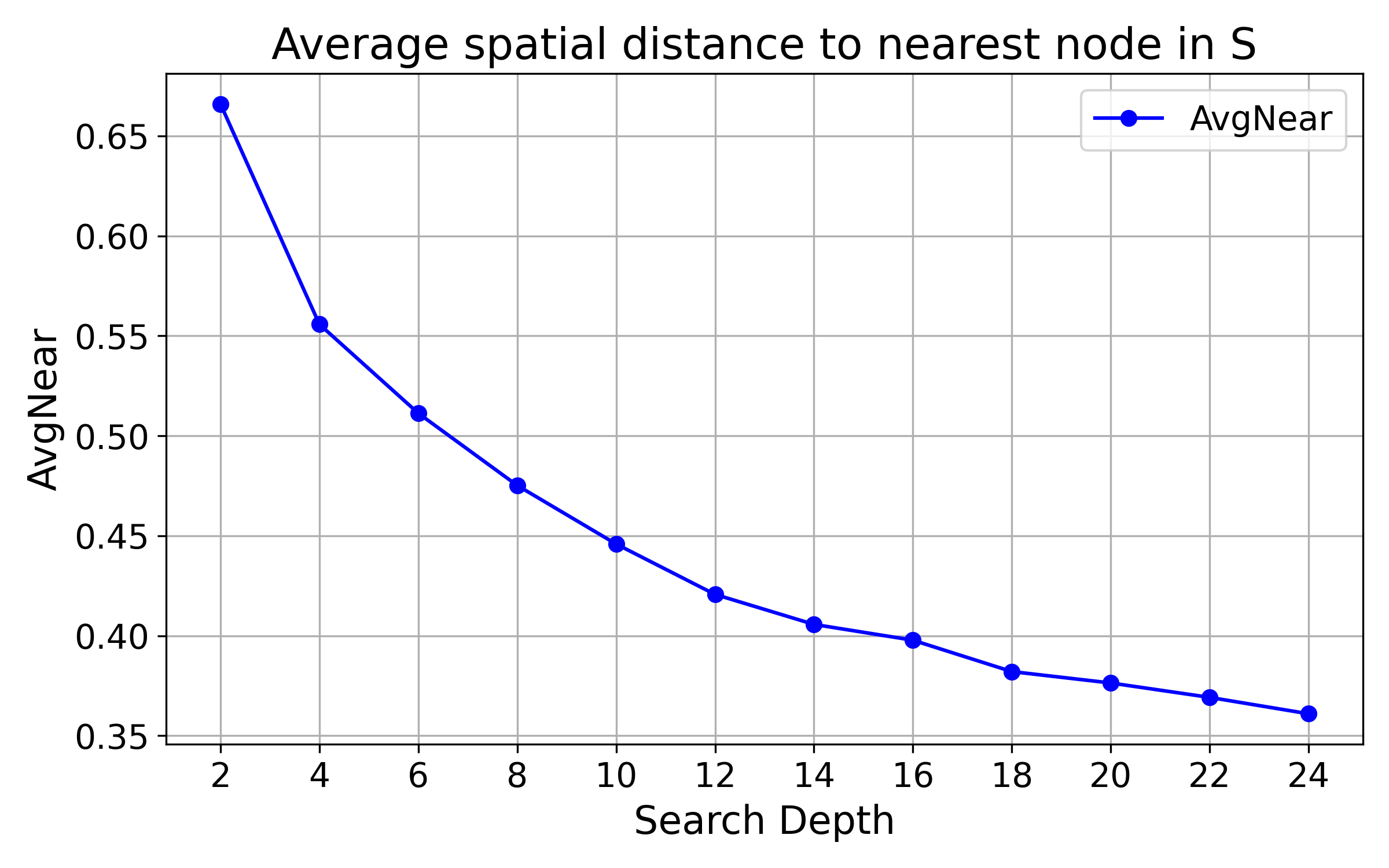}
        \caption{\scriptsize Nearest-Sample Distance}
    \end{subfigure}
    \hfill
    \begin{subfigure}[t]{0.245\textwidth}
        \centering
        \includegraphics[width=\textwidth,trim={0.2cm 0 0.2cm 1.04cm},clip]{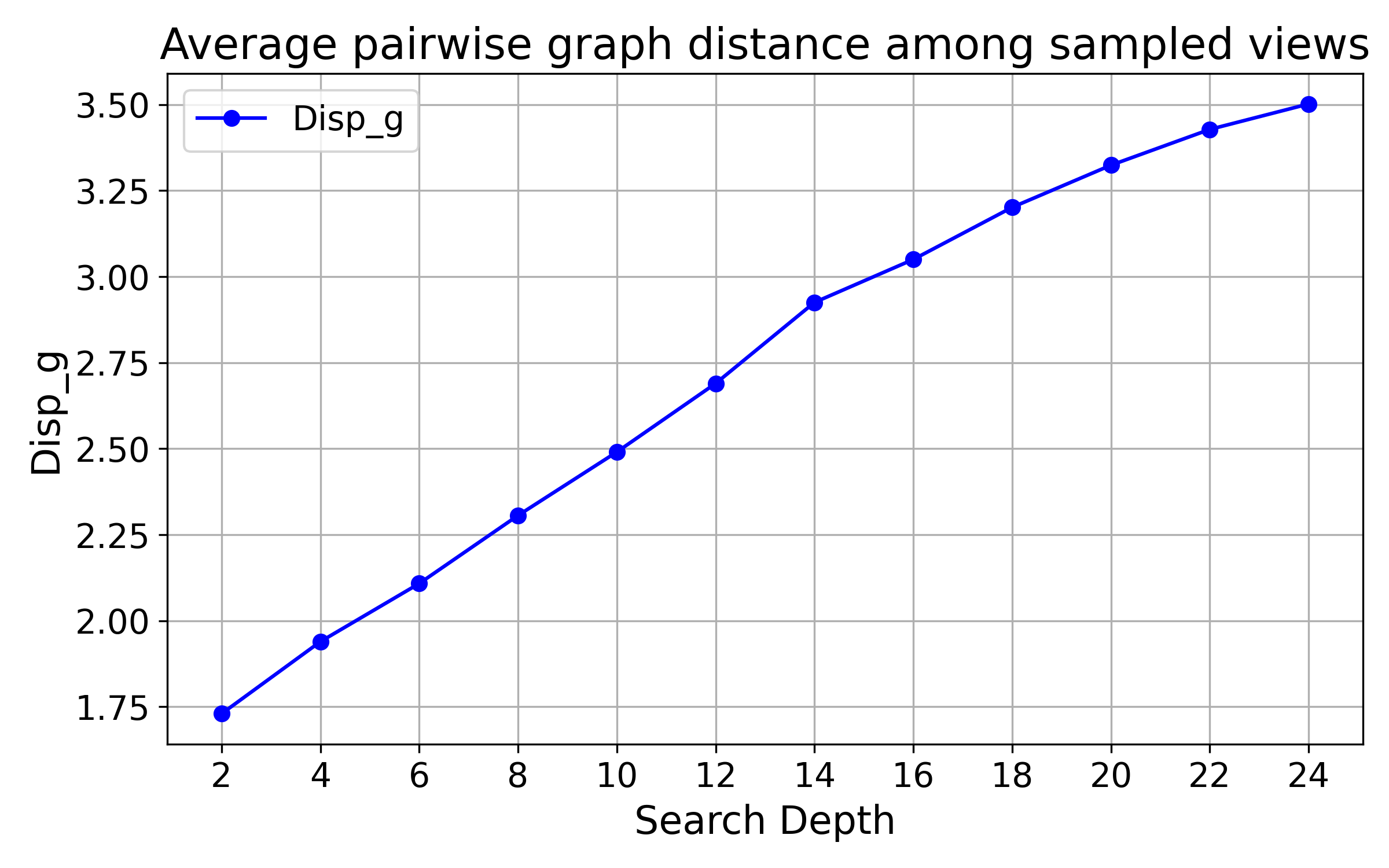}
        \caption{\scriptsize Graph Dispersion (pairwise hops)}
    \end{subfigure}
    \hfill
    \begin{subfigure}[t]{0.245\textwidth}
        \centering
        \includegraphics[width=\textwidth,trim={0.2cm 0 0.2cm 1.04cm},clip]{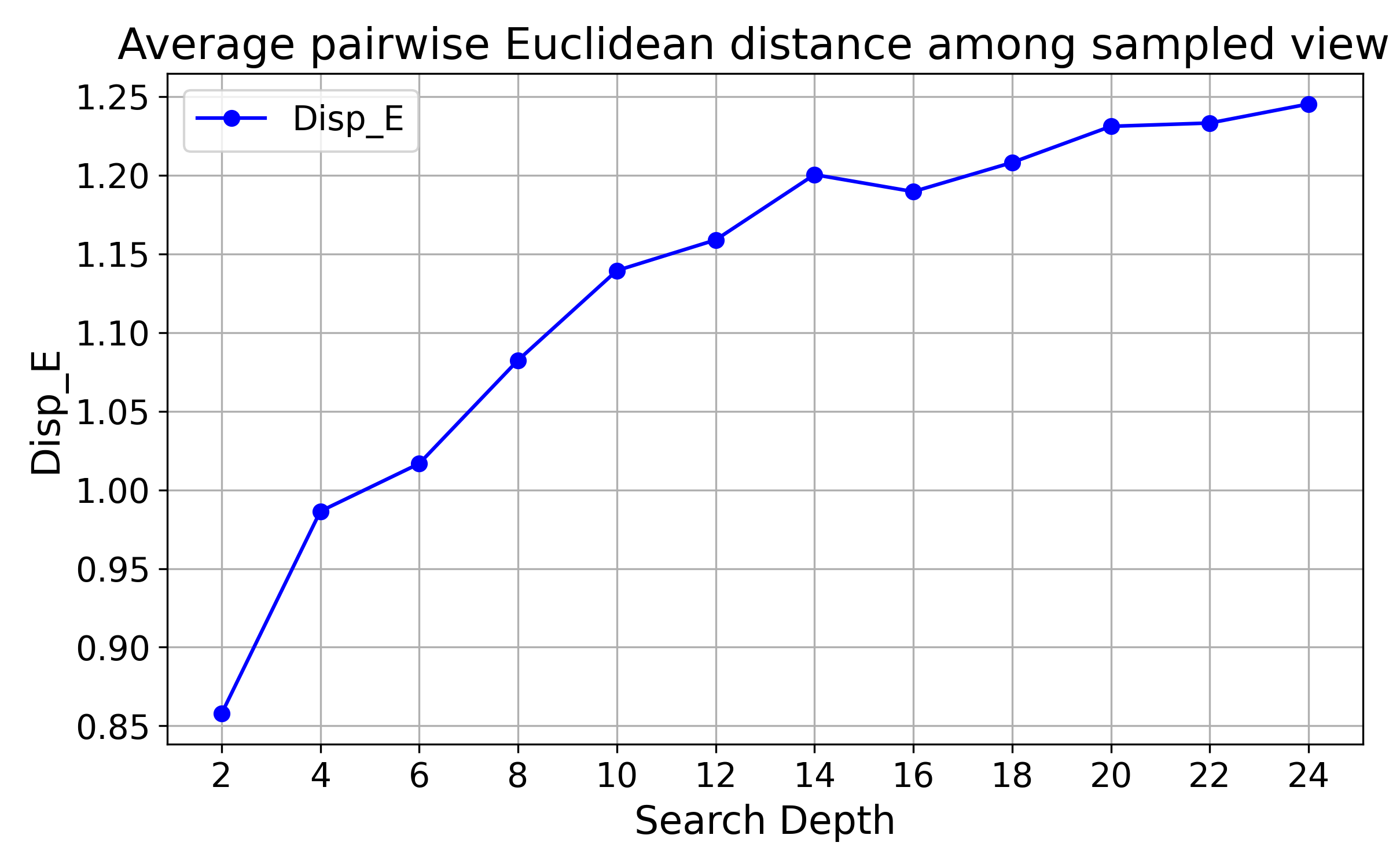}
        \caption{\scriptsize Euclidean Dispersion (pairwise distance)}
    \end{subfigure}
    
    \caption{\textbf{Coverage and sparsity vs.\ search depth.}
    Metrics in (a) and (b) evaluate coverage with respect to the \emph{full} view-graph, while (c) and (d) measure the sparsity of the \emph{sampled} subset. 
    As the search depth increases, the sampled set reaches a larger portion of the view-graph, as shown by the rise in $k$-hop (graph-distance) coverage in (a). 
    The average distance from each camera to its nearest sampled view decreases in (b), indicating broader spatial coverage. 
    At the same time, both graph dispersion (average pairwise graph distance) in (c) and Euclidean dispersion (average pairwise 3D distance) in (d) increase with depth, showing that the sampled views become more widely separated across the graph and in 3D space.}
    \label{fig:span_vs_depth}
\end{figure*}

To understand how greedy search depth $D$ affects the coverage and sparsity of the sampled views, we analyze several statistics on the view-graph.
Let $G$ denote the full view-graph of a scene and $S$ the set of sampled nodes.
The first two metrics quantify coverage with respect to the \emph{entire} graph $G$, while the last two measure sparsity \emph{within} the sampled subset $S$.

\medskip\noindent\textbf{k-hop graph coverage.} 
This metric measures how much of the view-graph is reached by the sampled views. Specifically, it computes the fraction of nodes in $G$ that lie within $k$ hops of any sampled node:
\begin{equation}
    \text{Cov}_{k}(G, S) = 
    \frac{1}{\lvert G \rvert}\lvert\{u\in G, v\in S \mid d_{G}(u, v)\le k \}\rvert,
\end{equation}
where $S$ is the subgraph of greedy sampled nodes and $d_{G}(u, v)$ is the shortest path from $u$ to $v$ on the graph $G$. A higher $\text{Cov}_{k}$ indicates broader topological coverage, i.e., the sampled set reaches many graph neighborhoods rather than remaining confined to a small region.

\medskip\noindent\textbf{Nearest-sample distance.}
To evaluate spatial coverage in 3D, we compute the average Euclidean distance from each camera to its closest sampled camera:
\begin{equation}
    \text{AvgNear}(G, S) = \frac{1}{\lvert G \rvert}\sum_{u\in G}\min_{v\in S}\|p_u - p_v\|_2,
\end{equation}
where $p_u$ and $p_v$ are camera positions.
Lower values mean the sampled views are spatially well-distributed and lie near many original cameras.

\medskip\noindent\textbf{Graph dispersion and Euclidean dispersion.}
To understand the sparsity of the sampled views, we calculate the average pairwise distance among sampled views(dispersion) based on graph distances and Euclidean distances:
\begin{align}
    \text{Disp}_\text{g}(S) & = \frac{1}{\lvert S \rvert(\lvert S \rvert-1)}\sum_{u,v \in S, u \neq v} d_{G}(u, v), \\ 
    \text{Disp}_\text{E}(S) & = \frac{1}{\lvert S \rvert(\lvert S \rvert-1)}\sum_{u,v \in S, u \neq v} \|p_u - p_v\|_2.
\end{align}
Higher dispersion values indicate that the sampled views are more sparsely distributed in both the graph and Euclidean space.

We compute these metrics for the top 100 scenes with the most registered images, evaluating 12 search depths and averaging over 8 sampling runs per depth. The number of sampled views is 24 for all samples.
Results are shown in Fig.~\ref{fig:span_vs_depth}, indicating that deeper searches yield higher coverage on the full graph (a,b) and produce sparser, more widely distributed sampled subsets (c,d).

\begin{table*}[t]
\centering
\small
\caption{Video Depth Estimation on Sintel~\cite{butler2012sintel}, Bonn~\cite{palazzolo2019bonn}, and KITTI~\cite{geiger2013kitti}. We report Absolute Relative Error (Abs Rel, lower is better) and the prediction accuracy at a threshold of $\delta\!<\!1.25$ (higher is better).}
\resizebox{0.8\linewidth}{!}{
\begin{tabular}{lccccccc}
\toprule
Method & Align &
\multicolumn{2}{c}{Sintel} & \multicolumn{2}{c}{Bonn} & \multicolumn{2}{c}{KITTI} \\
\cmidrule(lr){3-4} \cmidrule(lr){5-6} \cmidrule(lr){7-8}
 & & Abs Rel$\downarrow$ & $\delta<1.25\uparrow$ & Abs Rel$\downarrow$ & $\delta<1.25\uparrow$ & Abs Rel$\downarrow$ & $\delta<1.25\uparrow$ \\
\midrule
$\pi^3$                & \multirow{4}{*}{\textit{scale}}  
& 0.228 & 0.671 & 0.051 & 0.975 & \textbf{0.038} & \textbf{0.986} \\
$\pi^3$-\textsc{FT}    & 
& \textbf{0.213} & \textbf{0.713} & \textbf{0.047} & \textbf{0.978} & 0.040 & 0.985 \\
\cmidrule(lr){1-1} \cmidrule(lr){3-8}
VGGT                   &
& 0.294 & 0.649 & \textbf{0.055} & \textbf{0.971} & 0.072 & 0.965 \\
VGGT-\textsc{FT}       &
& \textbf{0.242} & \textbf{0.707} & 0.061 & 0.969 & \textbf{0.065} & \textbf{0.966} \\
\midrule
$\pi^3$                & \multirow{4}{*}{\textit{scale}\&\textit{shift}}
& 0.207 & 0.735 & 0.045 & 0.976 & \textbf{0.036} & \textbf{0.986} \\
$\pi^3$-\textsc{FT}    &
& \textbf{0.188} & \textbf{0.739} & \textbf{0.043} & \textbf{0.978} & 0.038 & 0.985 \\
\cmidrule(lr){1-1} \cmidrule(lr){3-8}
VGGT                   & 
& 0.226 & 0.683 & \textbf{0.049} & \textbf{0.974} & 0.059 & 0.961 \\
VGGT-\textsc{FT}       & 
& \textbf{0.197} & \textbf{0.728} & 0.056 & 0.973 & \textbf{0.056} & \textbf{0.964} \\
\bottomrule
\end{tabular}
}
\label{tab:videodepth}
\end{table*}

\begin{table*}[t]
\centering
\small
\caption{Monocular Depth Estimation on Sintel~\cite{butler2012sintel}, Bonn~\cite{palazzolo2019bonn}, KITTI~\cite{geiger2013kitti}, and NYU-v2~\cite{silberman2012nyuv2}.
We report Absolute Relative Error (Abs Rel, lower is better) and threshold accuracy $\delta\!<\!1.25$ (higher is better).}

\resizebox{0.8\linewidth}{!}{
\begin{tabular}{lcccccccc}
\toprule
Method &
\multicolumn{2}{c}{Sintel} & \multicolumn{2}{c}{Bonn} & \multicolumn{2}{c}{KITTI} & \multicolumn{2}{c}{NTU-v2} \\
\cmidrule(lr){2-3} \cmidrule(lr){4-5} \cmidrule(lr){6-7} \cmidrule(lr){8-9}
 & Abs Rel$\downarrow$ & $\delta<1.25\uparrow$ 
 & Abs Rel$\downarrow$ & $\delta<1.25\uparrow$ 
 & Abs Rel$\downarrow$ & $\delta<1.25\uparrow$ 
 & Abs Rel$\downarrow$ & $\delta<1.25\uparrow$ \\
\midrule
$\pi^3$ 
& \textbf{0.277} & 0.621 & 0.052 & 0.971 & 0.059 & \textbf{0.972} & 0.054 & 0.956 \\
$\pi^3$-\textsc{FT}
& 0.284 & \textbf{0.629} & \textbf{0.049} & \textbf{0.977} & \textbf{0.056} & \textbf{0.972} & \textbf{0.052} & \textbf{0.958} \\
\cmidrule(lr){1-9}
VGGT
& 0.331 & 0.600 & \textbf{0.051} & \textbf{0.974} & \textbf{0.089} & 0.939 & 0.055 & 0.953 \\
VGGT-\textsc{FT}
& \textbf{0.311} & \textbf{0.628} & 0.056 & \textbf{0.974} & 0.092 & \textbf{0.941} & \textbf{0.053} & \textbf{0.955} \\
\bottomrule
\end{tabular}
}
\label{tab:monodepth}
\end{table*}

\section{Training Details and Additional Results}
\subsection{Training Setup}

We finetune both $\pi^3$ and VGGT using their released pretrained checkpoints. All input images are first padded with white borders to a resolution of $518\times518$. During training, we apply random crops to these padded images, sampling aspect ratios uniformly from $[0.75, 1.0]$. We also apply random color jittering on training images. Each mini-batch contains up to 24 images drawn from \megascenesnew, with the number of views per batch randomly selected from $[2, 24]$. We process at most 96 images on each GPU. We also augment image orientations during training by randomly rotating images $90^\circ$ clockwise or counterclockwise with a probability of 0.2.

We use the original loss functions from $\pi^3$~\cite{wang2025pi3} and VGGT~\cite{wang2025vggt} to finetune the models. To preserve the geometric priors encoded in the pretrained models, we finetune only the Alternating-Attention modules, while keeping the point-cloud and camera decoders frozen. We further include BlendedMVS~\cite{yao2020blendedmvs} and TartanAir~\cite{wang2020tartanair} as additional training data for finetuning. Finetuning is performed for 100 epochs, where each epoch iterates over all scenes in the combined dataset. We use the AdamW optimizer with a peak learning rate of $1\times10^{-5}$, scheduled with linear warm-up followed by cosine annealing. All experiments are conducted on 4 NVIDIA A6000 GPUs.

\subsection{Additional Depth-Estimation Results}

We provide monocular and video depth results to complement the main paper. 
Following~\cite{wang2025cut3r,zhang2024monst3r,wang2025pi3}, we evaluate Absolute Relative Error (Abs Rel) 
and the accuracy at a threshold of $\delta < 1.25$. 
For monocular depth, we report performance on Sintel~\cite{butler2012sintel}, 
Bonn~\cite{palazzolo2019bonn}, KITTI~\cite{geiger2013kitti}, 
and NYU-v2~\cite{silberman2012nyuv2}. 
For video depth, we evaluate on Sintel~\cite{butler2012sintel}, Bonn~\cite{palazzolo2019bonn}, and KITTI~\cite{geiger2013kitti} under both 
\textit{scale} and \textit{scale\&shift} alignment settings. 
Our finetuned models maintain competitive performance across all datasets, 
demonstrating that the adaptation to in-the-wild imagery does not degrade 
their depth-estimation ability.

\subsection{Results on Doppelganger Scenes}

\begin{figure}[t]
    \centering
    \includegraphics[width=\linewidth]{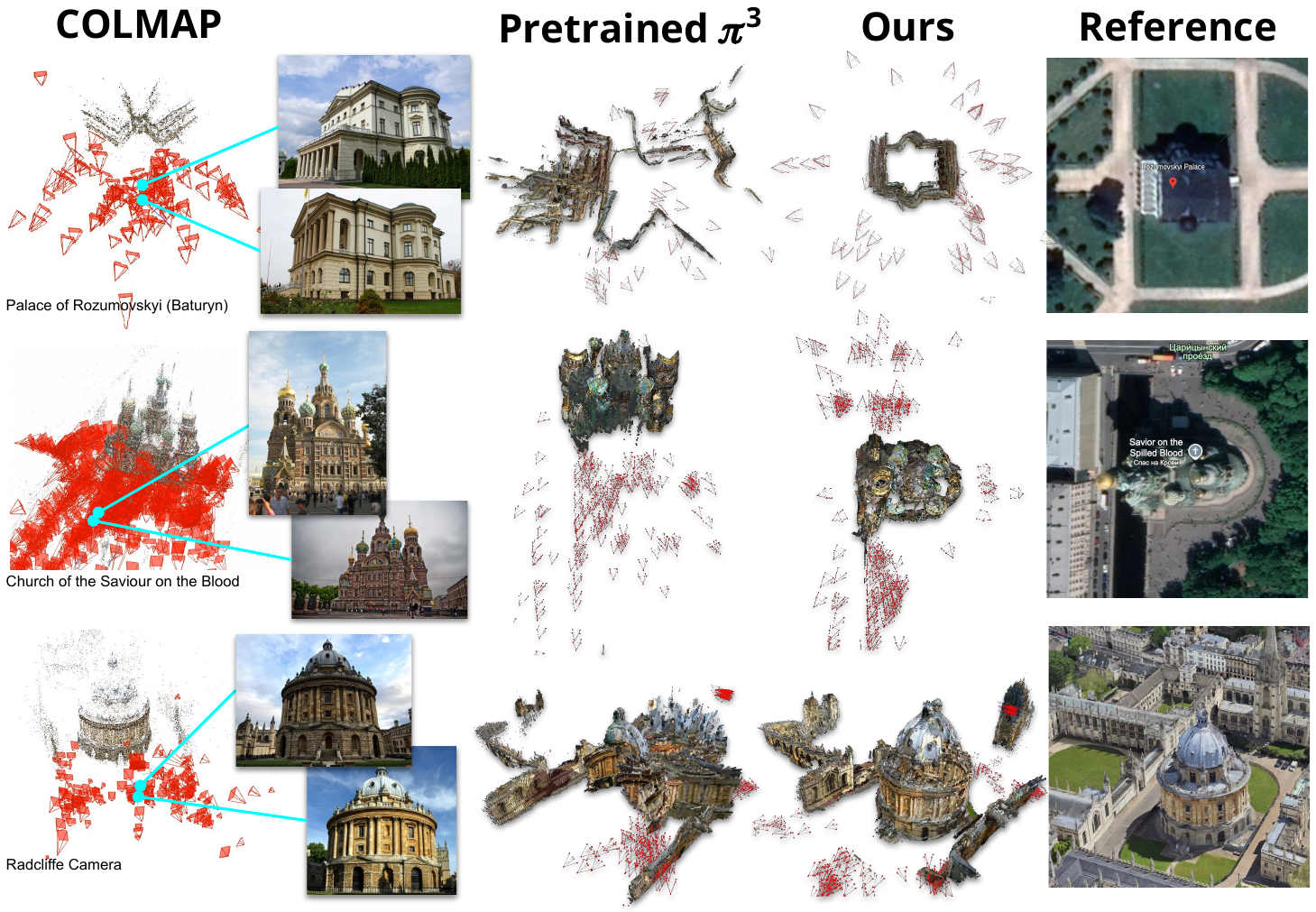}
    \vspace{-20pt}
    \caption{\footnotesize\textbf{Disambiguation of doppelganger scenes.}
Each example shows a pair of visually similar structures that cause classical SfM (COLMAP) and pretrained $\pi^3$ to collapse into incorrect or merged reconstructions. 
In contrast, our finetuned model correctly distinguishes the symmetric or repetitive sides of the same building, reconstructing consistent geometry for each viewpoint. 
Reference views from Google Earth are provided for comparison, confirming that our model resolves these ambiguities and recovers accurate global structure under challenging visual similarity.
}

    \label{fig:dopp_vis}
    \vspace{-3mm}
\end{figure}

Doppelganger cases often cause both classical SfM pipelines and pretrained feed-forward models to fail, merging distinct structures into a single incorrect reconstruction. 
As shown in Fig.~\ref{fig:dopp_vis}, our fine-tuned $\pi^3$ model correctly distinguishes visually similar but distinct structures within each landmark and recovers geometry consistent with reference aerial imagery, indicating improved reconstruction of global scene layout.

\begin{figure}[t]
    \centering
    \includegraphics[width=1\linewidth]{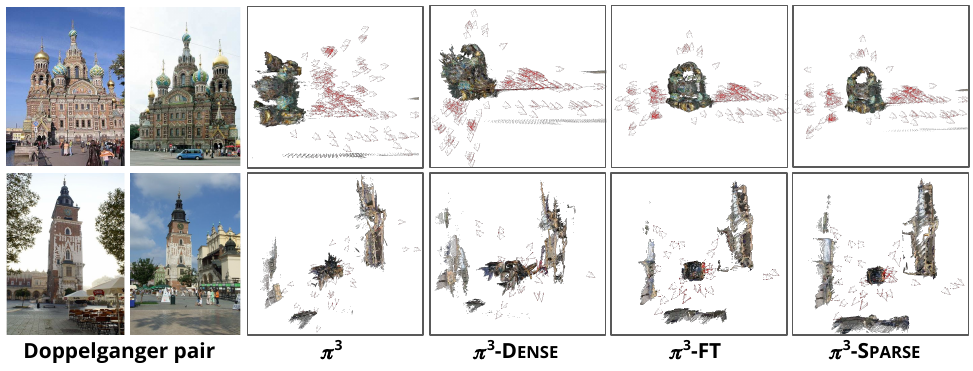}
    \caption{\footnotesize\textbf{Comparison of ablated models on doppelganger scenes} We show predictions from the pre-trained model and ablated models on two doppelganger scenes. Disambiguation behavior holds across fine-tuned variants with sparsity-aware sampling, while the pre-trained model and model finetuned with densely sampled views are less robust to doppelgangers.}
    \label{fig:dopp_ablation}

\end{figure}

To evaluate the effectiveness of different sampling strategies on doppelganger scenes, we evaluate the pretrained $\pi^3$ and finetuned $\pi^3$ on doppelganger scenes and show results in fig.\ref{fig:dopp_ablation}. Results indicate that pretrained models and dense-only fine-tuning are less robust to ambiguity, while finetuning with sparsity-aware sampling (e.g., mixed or sparse) tends to improve disambiguation, suggesting sparsity-aware sampling helps.

\subsection{Quantitative results on Long-tail scenes}

\begin{figure*}[t]
    \centering
    \includegraphics[width=1\linewidth]{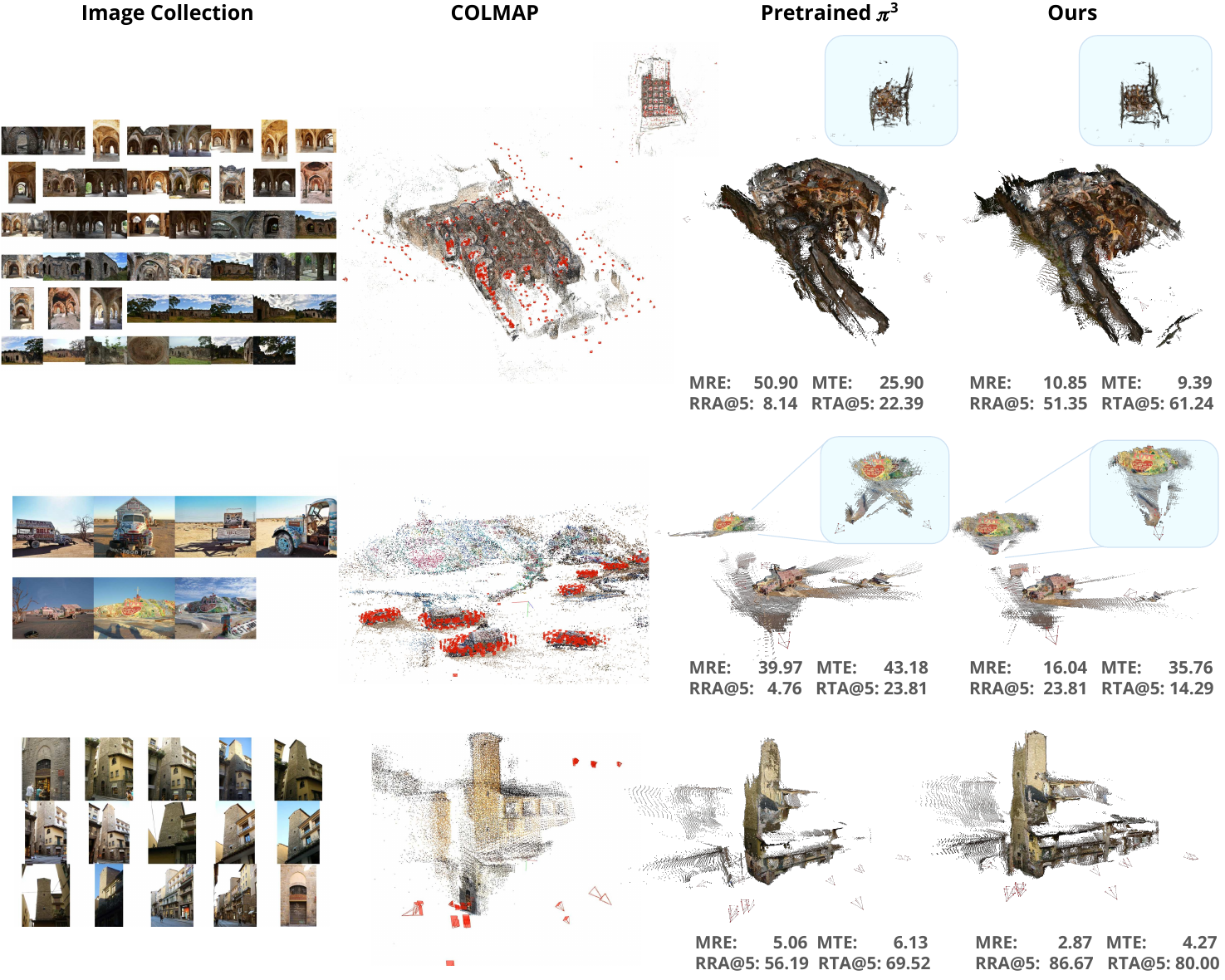}
    \vspace{-20pt}
    \caption{\footnotesize\textbf{Quantitative results on Long-tail scenes.} Our model performs better on scenes with strong ambiguities (first row) and on scenes with minimal overlap across different scene components (second row). For a more densely photographed scene that still exhibits large viewpoint variation (third row), our model not only reduces pose error but also reconstructs a more complete point cloud.}
    \label{fig:openheritage3d}
    \vspace{-3mm}
\end{figure*}

To enable quantitative evaluation on long-tail scenes, we augment MegaScenes with additional observations from external cultural heritage datasets~\cite{chiabrando2023salvationmountain,cyark2020greatmosque,richter2023torredeibaldovinetti} and jointly register all images using COLMAP. The quantitative and qualitative results of this long-tail evaluation are shown in Fig.~\ref{fig:openheritage3d}. Our model consistently reduces the mean relative rotation and translation errors across all scenes, while also producing more complete point clouds.

\subsection{Limitations}

\begin{figure*}[t]
    \centering
    \includegraphics[width=1\linewidth]{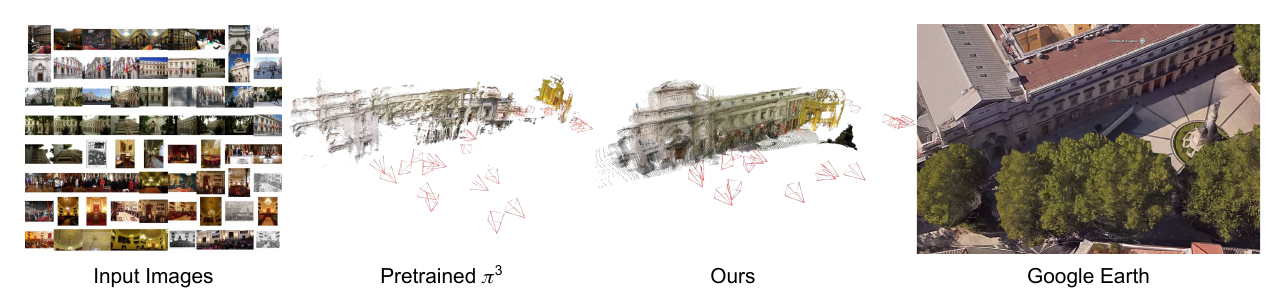}
    \caption{\textbf{Limitations.}
    This example contains images from two disjoint parts of the scene: indoor photos with warm lighting (producing a yellowish point cloud) and outdoor photos (producing a white point cloud). Pretrained $\pi^3$ struggles to handle such mixed inputs and produces inconsistent geometry. Our finetuned model is more robust in this setting, but both models still fuse the indoor and outdoor structures into a single reconstruction without separating them.}
    \label{fig:limitation}
    \vspace{-3mm}
\end{figure*}

Long-tail scenes often contain fragmented viewpoints, where different subsets of images capture disjoint parts of the scene (e.g., indoor and outdoor areas) without overlapping views to connect them. 
When such mixed collections
are fed into the models 
at once,
both pretrained and finetuned $\pi^3$ may blend these unrelated regions into a single 3D structure,
as illustrated in Fig.\ref{fig:limitation}. 
While our finetuned model handles these mixtures more robustly than the pretrained baseline, enabling the model to reason robustly about disconnected components and produce reasonable overall layouts still remains a challenge.

\end{document}